\documentclass[11pt]{article}
\usepackage{iftex}

\usepackage[preprint]{colm2026_conference}

\ifPDFTeX
\usepackage[T2A,T1]{fontenc}

\usepackage[utf8]{inputenc}
\fi

\usepackage{microtype} 

\usepackage{graphicx}
\usepackage{url}
\usepackage{hyperref}
\usepackage{xcolor}
\usepackage{mathtools}
\usepackage{physics}
\usepackage{cleveref}
\usepackage{booktabs}
\usepackage{tabularx}
\usepackage{multirow}
\usepackage{siunitx}
\usepackage{multicol}
\usepackage{listings}
\usepackage{enumitem}

\usepackage{expex}

\let\tok\texttt
\def\NP{\textrm{NP}}
\def\VP{\textrm{VP}}
\def\VB{\textrm{VB}}

\DeclareMathOperator{\BLEU}{\textrm{BLEU}}
\DeclareMathOperator{\BP}{\textrm{BP}}
\DeclareMathOperator{\chrFPP}{\textrm{chrF\textsuperscript{++}}}
\DeclareMathOperator{\chrP}{\textrm{chrP}}
\DeclareMathOperator{\chrR}{\textrm{chrR}}

\usepackage{tcolorbox}
\tcbuselibrary{skins,breakable}
\definecolor{sky}{HTML}{79c8fc}
\definecolor{palesky}{HTML}{e6f5ff}

\definecolor{goldenrod}{HTML}{fdedd7}

\definecolor{paper}{HTML}{FFFCF0}
\definecolor{base50}{HTML}{F2F0E5}
\definecolor{base100}{HTML}{E6E4D9}
\definecolor{base150}{HTML}{DAD8CE}
\definecolor{base200}{HTML}{CECDC3}
\definecolor{base300}{HTML}{B7B5AC}
\definecolor{base400}{HTML}{9F9D96}
\definecolor{base500}{HTML}{878580}
\definecolor{base600}{HTML}{6F6E69}
\definecolor{base700}{HTML}{575653}
\definecolor{base800}{HTML}{403E3C}
\definecolor{base850}{HTML}{343331}
\definecolor{base900}{HTML}{282726}
\definecolor{base950}{HTML}{1C1B1A}
\definecolor{black}{HTML}{100F0F}

\definecolor{canvas}{HTML}{FBF8F4}

\definecolor{canvasOrange}{HTML}{FBF8F4}  
\definecolor{canvasRed}{HTML}{F9F6F7}
\definecolor{canvasYellow}{HTML}{F9F7F6}
\definecolor{canvasGreen}{HTML}{F6F9F6}
\definecolor{canvasCyan}{HTML}{F6F9F8}
\definecolor{canvasBlue}{HTML}{F6F8F9}
\definecolor{canvasPurple}{HTML}{F8F6F9}
\definecolor{canvasMagenta}{HTML}{F9F6F7}

\definecolor{red}{HTML}{C84049}
\definecolor{red50}{HTML}{F7EEEE}
\definecolor{red100}{HTML}{EEDDDE}
\definecolor{red150}{HTML}{E7CBCD}
\definecolor{red200}{HTML}{DFB9BC}
\definecolor{red300}{HTML}{D19498}
\definecolor{red400}{HTML}{C56D73}
\definecolor{red500}{HTML}{C84049}  
\definecolor{red600}{HTML}{B3323B}
\definecolor{red700}{HTML}{8D252C}
\definecolor{red800}{HTML}{661A1F}
\definecolor{red900}{HTML}{3E0F12}
\definecolor{red950}{HTML}{290A0C}

\definecolor{orange}{HTML}{D65F3D}
\definecolor{orange50}{HTML}{F7EFED}
\definecolor{orange100}{HTML}{F0E0DB}
\definecolor{orange150}{HTML}{E9D0C8}
\definecolor{orange200}{HTML}{E3BFB5}
\definecolor{orange300}{HTML}{D79E8E}
\definecolor{orange400}{HTML}{CD7C65}
\definecolor{orange500}{HTML}{D65F3D}  
\definecolor{orange600}{HTML}{BF4826}
\definecolor{orange700}{HTML}{97371C}
\definecolor{orange800}{HTML}{6D2713}
\definecolor{orange900}{HTML}{42170A}
\definecolor{orange950}{HTML}{2C0F07}

\definecolor{yellow}{HTML}{E5974D}
\definecolor{yellow50}{HTML}{F9F2EC}
\definecolor{yellow100}{HTML}{F3E5D8}
\definecolor{yellow150}{HTML}{EED8C4}
\definecolor{yellow200}{HTML}{E9CBAF}
\definecolor{yellow300}{HTML}{E0B185}
\definecolor{yellow400}{HTML}{E9AC72}
\definecolor{yellow500}{HTML}{E5974D}  
\definecolor{yellow600}{HTML}{CC7019}
\definecolor{yellow700}{HTML}{A15712}
\definecolor{yellow800}{HTML}{743E0B}
\definecolor{yellow900}{HTML}{472506}
\definecolor{yellow950}{HTML}{2F1904}

\definecolor{green}{HTML}{589358}
\definecolor{green50}{HTML}{F0F4F0}
\definecolor{green100}{HTML}{E2E9E2}
\definecolor{green150}{HTML}{D3DFD3}
\definecolor{green200}{HTML}{C4D4C4}
\definecolor{green300}{HTML}{A6BFA6}
\definecolor{green400}{HTML}{86AC86}
\definecolor{green500}{HTML}{589358}  
\definecolor{green600}{HTML}{569056}
\definecolor{green700}{HTML}{427042}
\definecolor{green800}{HTML}{2F512F}
\definecolor{green900}{HTML}{1C311C}
\definecolor{green950}{HTML}{122112}

\definecolor{cyan}{HTML}{4ECDC4}
\definecolor{cyan50}{HTML}{EEF7F6}
\definecolor{cyan100}{HTML}{DCEFEE}
\definecolor{cyan150}{HTML}{CAE7E5}
\definecolor{cyan200}{HTML}{B8E0DD}
\definecolor{cyan300}{HTML}{93D2CE}
\definecolor{cyan400}{HTML}{6BC7C0}
\definecolor{cyan500}{HTML}{4ECDC4}  
\definecolor{cyan600}{HTML}{31B5AC}
\definecolor{cyan700}{HTML}{248E87}
\definecolor{cyan800}{HTML}{196761}
\definecolor{cyan900}{HTML}{0E3E3B}
\definecolor{cyan950}{HTML}{092A27}

\definecolor{blue}{HTML}{2684BA}
\definecolor{blue50}{HTML}{EDF4F7}
\definecolor{blue100}{HTML}{DCE8EF}
\definecolor{blue150}{HTML}{CADDE8}
\definecolor{blue200}{HTML}{B7D2E1}
\definecolor{blue300}{HTML}{91BCD4}
\definecolor{blue400}{HTML}{69A6C9}
\definecolor{blue500}{HTML}{2684BA}  
\definecolor{blue600}{HTML}{3484B1}
\definecolor{blue700}{HTML}{1D6A96}
\definecolor{blue800}{HTML}{134C6C}
\definecolor{blue900}{HTML}{0B2E42}
\definecolor{blue950}{HTML}{071F2C}

\definecolor{purple}{HTML}{9966A2}
\definecolor{purple50}{HTML}{F4F0F4}
\definecolor{purple100}{HTML}{E8E2E9}
\definecolor{purple150}{HTML}{DDD3DF}
\definecolor{purple200}{HTML}{D2C4D4}
\definecolor{purple300}{HTML}{BCA5C0}
\definecolor{purple400}{HTML}{A686AC}
\definecolor{purple500}{HTML}{9966A2}  
\definecolor{purple600}{HTML}{87568F}
\definecolor{purple700}{HTML}{694270}
\definecolor{purple800}{HTML}{4C2F51}
\definecolor{purple900}{HTML}{2E1C31}
\definecolor{purple950}{HTML}{1E1320}

\definecolor{magenta}{HTML}{C75B7A}
\definecolor{magenta50}{HTML}{F6EEF1}
\definecolor{magenta100}{HTML}{EEDDE2}
\definecolor{magenta150}{HTML}{E6CCD3}
\definecolor{magenta200}{HTML}{DEBAC5}
\definecolor{magenta300}{HTML}{CF96A6}
\definecolor{magenta400}{HTML}{C27088}
\definecolor{magenta500}{HTML}{C75B7A}  
\definecolor{magenta600}{HTML}{AD385A}
\definecolor{magenta700}{HTML}{882B45}
\definecolor{magenta800}{HTML}{621D31}
\definecolor{magenta900}{HTML}{3B111D}
\definecolor{magenta950}{HTML}{280B13}

\colorlet{persimmon}{orange500}
\colorlet{turquoise}{green600}

\newcommand{\suff}[1]{\textcolor{persimmon}{#1}}
\newcommand{\pron}[1]{\textcolor{turquoise}{#1}}
\newcommand{\fade}[1]{\textcolor{base500}{#1}}

\newcommand{\subj}[1]{\textcolor{red500}{#1}}
\newcommand{\verbal}[1]{\textcolor{cyan600}{#1}}
\newcommand{\obj}[1]{\textcolor{yellow600}{#1}}

\newtcolorbox{user}[1][]{
    enhanced,
    after skip=8mm,
    title=#1,
    breakable = true,
    fonttitle=\small\sffamily\scshape\bfseries,
    fontupper=\rmfamily\small,
    coltitle=persimmon,
    colbacktitle=canvas,
    boxrule=0.5pt,
    colframe=persimmon,
    parbox=false,
    left=3pt,
    right=3pt,
    overlay={%
        \ifcase\tcbsegmentstate
        \or%
        \else%
        \fi%
    },
    colback = goldenrod,
}

\usepackage{tcolorbox}
\usepackage{tikz}
\usepackage{xcolor}

\definecolor{canvas}{HTML}{fbf8f4}
\definecolor{goldenrod}{HTML}{fdedd7}
\definecolor{persimmon}{HTML}{d65f3d}
\definecolor{mygreen}{RGB}{0,128,0} 
\definecolor{myred}{RGB}{204,0,0}   

\newcounter{paranum}
\newcounter{boxpara}
\newbool{tcb@numberparas}
\boolfalse{tcb@numberparas}

\tcbuselibrary{most}
\usetikzlibrary{decorations.pathmorphing}

\makeatletter
\newdimen\chat@zig@amp
\newdimen\chat@zig@seg
\newdimen\chat@zig@x
\newdimen\chat@zig@remaining
\newdimen\chat@zig@step
\newdimen\chat@zig@half
\newdimen\chat@xa
\newdimen\chat@xb
\newdimen\chat@ya
\newcount\chat@zig@count
\newbox\chattitleboxsaved
\newcommand{\chattitlebox}{\copy\tcb@titlebox}

\newcommand{\chatzigzagline}[2]{%
  \pgfextractx{\chat@xa}{#1}%
  \pgfextractx{\chat@xb}{#2}%
  \pgfextracty{\chat@ya}{#1}%
  \chat@zig@x=\chat@xa%
  \chat@zig@remaining=\chat@xb%
  \advance\chat@zig@remaining by -\chat@xa%
  \pgfmathsetmacro{\chat@zig@width}{abs((\chat@zig@remaining)/1pt)}%
  \pgfmathsetmacro{\chat@zig@segpt}{\chat@zig@seg/1pt}%
  \pgfmathtruncatemacro{\chat@zig@n}{max(1,round(\chat@zig@width/\chat@zig@segpt))}%
  \chat@zig@count=\chat@zig@n\relax%
  \chat@zig@step=\chat@zig@remaining%
  \ifnum\chat@zig@count>0\relax%
    \divide\chat@zig@step by \chat@zig@count%
  \fi%
  \chat@zig@half=\chat@zig@step%
  \divide\chat@zig@half by 2%
  \pgfpathlineto{\pgfqpoint{\chat@zig@x}{\chat@ya}}%
  \chat@zig@count=0%
  \loop%
    \ifnum\chat@zig@count<\chat@zig@n\relax%
      \advance\chat@zig@x by \chat@zig@half%
      \advance\chat@zig@count by 1%
      \ifodd\chat@zig@count%
        \pgfpathlineto{\pgfqpoint{\chat@zig@x}{\dimexpr\chat@ya+\chat@zig@amp\relax}}%
      \else%
        \pgfpathlineto{\pgfqpoint{\chat@zig@x}{\dimexpr\chat@ya-\chat@zig@amp\relax}}%
      \fi%
      \advance\chat@zig@x by \chat@zig@half%
      \pgfpathlineto{\pgfqpoint{\chat@zig@x}{\chat@ya}}%
  \repeat%
}
\makeatother

\AddToHook{para/begin}{%
  \ifbool{tcb@numberparas}{%
    \stepcounter{boxpara}%
    \vadjust{%
      \smash{%
        \makebox[\linewidth][r]{%
          \scriptsize\color{persimmon}%
          \textpilcrow~\arabic{boxpara}\hspace{0.4em}%
        }%
      }%
    }%
  }{}%
}

\tcbset{
  chatbox/.style={
    enhanced,
    boxrule=0.9pt,
    arc=2pt,
    colback=white,
    left=3mm,right=3mm,top=4mm,bottom=3mm,
    before skip=10pt,
    after skip=10pt,
    breakable = true,
    frame hidden,
    interior hidden,
    underlay boxed title={%
      \global\setbox\chattitleboxsaved=\copy\csname tcb@titlebox\endcsname%
      \global\setbox\csname tcb@titlebox\endcsname=\hbox{}%
    },
    boxed title style={
      arc=5pt,
      boxrule=0.9pt,
      left=1.3mm,
      right=1.3mm,
      top=0.6mm,
      bottom=0.6mm,
    },
    overlay unbroken={},
    overlay first={},
    overlay unbroken app={%
      \pgftext[at={\pgfpointanchor{title}{center}}]{\box\chattitleboxsaved}%
    },
    overlay first app={%
      \pgftext[at={\pgfpointanchor{title}{center}}]{\box\chattitleboxsaved}%
    },
    overlay middle={},
    overlay last={},
    fontupper=\rmfamily\small,
    fonttitle=\scriptsize\sffamily\bfseries,
    attach boxed title to top left={yshift=-3mm,xshift=4mm},
    finish={},
    no underlay unbroken,
    no underlay first,
    no underlay middle,
    no underlay last,
    underlay unbroken={%
      \begin{pgfscope}%
      \pgfsetcornersarced{\pgfqpoint{2pt}{2pt}}%
      \pgfpathmoveto{\pgfpointlineattime{0.02}{\pgfpointanchor{frame}{north west}}{\pgfpointanchor{frame}{south west}}}%
      \pgfpathlineto{\pgfpointanchor{frame}{north west}}%
      \pgfpathlineto{\pgfpointanchor{frame}{north east}}%
      \pgfpathlineto{\pgfpointanchor{frame}{south east}}%
      \pgfpathlineto{\pgfpointanchor{frame}{south west}}%
      \pgfpathclose%
      \pgfsetfillcolor{tcbcolback}%
      \pgfsetfillopacity{1}%
      \pgfsetstrokecolor{tcbcolframe}%
      \pgfsetstrokeopacity{1}%
      \pgfsetlinewidth{0.9pt}%
      \pgfusepath{fill,stroke}%
      \end{pgfscope}%
    },
    underlay first={%
      \begin{pgfscope}%
      \pgfsetcornersarced{\pgfqpoint{2pt}{2pt}}%
      \pgfpathmoveto{\pgfpointlineattime{0.02}{\pgfpointanchor{frame}{north west}}{\pgfpointanchor{frame}{south west}}}%
      \pgfpathlineto{\pgfpointanchor{frame}{north west}}%
      \pgfpathlineto{\pgfpointanchor{frame}{north east}}%
      \pgfpathlineto{\pgfpointanchor{frame}{south east}}%
      \pgfsetcornersarced{\pgfqpoint{0pt}{0pt}}%
      \chatzigzagline{\pgfpointanchor{frame}{south east}}{\pgfpointanchor{frame}{south west}}%
      \pgfpathclose%
      \pgfsetfillcolor{tcbcolback}%
      \pgfsetfillopacity{1}%
      \pgfsetstrokecolor{tcbcolframe}%
      \pgfsetstrokeopacity{1}%
      \pgfsetlinewidth{0.9pt}%
      \pgfusepath{fill,stroke}%
      \end{pgfscope}%
    },
    underlay middle={%
      \begin{pgfscope}%
      \pgfsetcornersarced{\pgfqpoint{0pt}{0pt}}%
      \pgfpathmoveto{\pgfpointanchor{frame}{north west}}%
      \chatzigzagline{\pgfpointanchor{frame}{north west}}{\pgfpointanchor{frame}{north east}}%
      \pgfpathlineto{\pgfpointanchor{frame}{south east}}%
      \chatzigzagline{\pgfpointanchor{frame}{south east}}{\pgfpointanchor{frame}{south west}}%
      \pgfpathclose%
      \pgfsetfillcolor{tcbcolback}%
      \pgfsetfillopacity{1}%
      \pgfsetstrokecolor{tcbcolframe}%
      \pgfsetstrokeopacity{1}%
      \pgfsetlinewidth{0.9pt}%
      \pgfusepath{fill,stroke}%
      \end{pgfscope}%
    },
    underlay last={%
      \begin{pgfscope}%
      \pgfsetcornersarced{\pgfqpoint{0pt}{0pt}}%
      \pgfpathmoveto{\pgfpointanchor{frame}{north west}}%
      \chatzigzagline{\pgfpointanchor{frame}{north west}}{\pgfpointanchor{frame}{north east}}%
      \pgfsetcornersarced{\pgfqpoint{2pt}{2pt}}%
      \pgfpathlineto{\pgfpointanchor{frame}{south east}}%
      \pgfpathlineto{\pgfpointanchor{frame}{south west}}%
      \pgfpathclose%
      \pgfsetfillcolor{tcbcolback}%
      \pgfsetfillopacity{1}%
      \pgfsetstrokecolor{tcbcolframe}%
      \pgfsetstrokeopacity{1}%
      \pgfsetlinewidth{0.9pt}%
      \pgfusepath{fill,stroke}%
      \end{pgfscope}%
    },
    before upper={%
      \setlength{\parskip}{0.5\baselineskip}%
    },
  },
  numbered/.style={
    left=8mm,
    before upper={%
      \setlength{\parskip}{0.5\baselineskip}%
      \setcounter{paranum}{0}%
      \def\paranumlabel{\refstepcounter{paranum}\llap{\textcolor{canvas!40!black}{\texttt{\textpilcrow\theparanum}}\hspace{2mm}}}%
      \everypar{\paranumlabel}%
      \let\oldpar\par%
      \def\par{\ifhmode\vadjust{\vskip0.5\baselineskip}\fi\oldpar}%
    },
  },
  linenumbered/.style={
    left=6mm,
    before upper={%
      \setlength{\parskip}{0.5\baselineskip}%
      \internallinenumbers%
      \resetlinenumber%
      \modulolinenumbers[5]%
      \linenumbersep=2mm%

    },
  },
}

\newtcolorbox[auto counter]{prompt}[1][]{%
  chatbox,
  title={\lsstyle\uppercase{prompt~\thetcbcounter}},
  coltitle=persimmon,
  boxed title style={
    colback=canvas,
    colframe=persimmon,
  },
  colframe=persimmon,
  colback=goldenrod!70!white,
  attach boxed title to top right={yshift=-3mm,xshift=-4mm},
  #1
}

\newtcolorbox[auto counter,use counter from=prompt]{llm}[1][]{%
  chatbox,
  title={\lsstyle\uppercase{llm response~\thetcbcounter}},
  coltitle=persimmon,
  boxed title style={
    colback=canvas,
    colframe=persimmon,
  },
  colframe=canvas!80!black,
  colback=canvas!70!white,
  #1
}

\lstdefinestyle{grammartt}{
  basicstyle=\ttfamily\small,
  columns=fullflexible,
  keepspaces=true,
  breaklines=true,
  breakatwhitespace=false,
  showstringspaces=false,
  frame=none
}

\usepackage{noto-sans}        

\newlist{compactenum}{enumerate}{4}
\setlist[compactenum,1]{nolistsep,label=\arabic*.,leftmargin=1.5em}

\title{Evaluating In-Context Translation with Synchronous Context-Free Grammar Transduction}

\author{Jackson Petty$^\lambda$, Jaulie Goe$^\delta$ \& Tal Linzen$^{\lambda\delta}$ \\
Department of Linguistics$^\lambda$ and Center for Data Science$^\delta$\\
New York University\\
\texttt{\{petty,jg5059,linzen\}@nyu.edu}
}

\begin{document}

\maketitle

\begin{abstract}
Low-resource languages pose a challenge for machine translation with large language models (LLMs), which require large amounts of training data. One potential way to circumvent this data dependence is to rely on LLMs' ability to use in-context descriptions of languages, like textbooks and dictionaries. To do so, LLMs must be able to infer the link between the languages' grammatical descriptions and the sentences in question. Here we isolate this skill using a formal analogue of the task: string transduction based on a formal grammar provided in-context. We construct synchronous context-free grammars which define pairs of formal languages designed to model particular aspects of natural language grammar, morphology, and written representation. Using these grammars, we measure how well LLMs can translate sentences from one formal language into another when given both the grammar and the source-language sentence. We vary the size of the grammar, the lengths of the sentences, the syntactic and morphological properties of the languages, and their written script. We note three key findings. First, LLMs' translation accuracy decreases markedly as a function of grammar size and sentence length. Second, differences in morphology and written representation between the source and target languages can strongly diminish model performance. Third, we examine the types of errors committed by models and find they are most prone to recall the wrong words from the target language vocabulary, hallucinate new words, or leave source-language words untranslated.
\end{abstract}

\section{Introduction}

Many of the world's languages lack enough written data for training neural machine translation systems, which depend heavily on large parallel and monolingual corpora \citep{kim-2020-when, zhu-2024-multilingual,nllbTeam-2024-scaling,alves-2024-tower,dang-2024-aya,ataman-2025-machine,omnilingualMtTeam-2026-omnilingual}. LLMs' increasingly sophisticated ability to reference and manipulate information provided in-context \citep{brown-2020-language-b,wei-2023-chain-of-thought, wei-2022-emergent, vodrahalli-2024-michelangelo} suggests a potential solution to this problem: could a language model translate into a language it has not been trained on by making use of descriptions of that language, such as textbooks, reference grammars, and dictionaries, provided in-context at inference time? Such in-context approaches would not only mitigate the problem of data scarcity facing low-resource languages but also allow translation systems themselves to be more adaptable to new languages and domains without needing additional or specialized training. 

Recent work evaluates LLMs in this setting, which we refer to as \textbf{in-context machine translation} (ICMT). \citet{tanzer-2023-benchmark} and \citet{geminiTeam-2024-gemini} study how well LLMs can translate into Kalamang, an Indonesian language with very little written corpora, on the basis of a reference grammar and dictionary. They find they fare decently well against a baseline of translations from non-native speakers who have access to similar resources. Yet the ability of LLMs to use grammatical descriptions for ICMT has been questioned by \citet{aycock-2024-llms}, who find that models rely primarily on example translations rather than the grammatical descriptions themselves, suggesting that reference grammars and dictionaries alone cannot substitute for parallel corpora in low-resource settings. Evaluating LLMs' ICMT performance is further complicated by the fact that, because sentences can typically be translated in many different ways, machine translation typically needs to be evaluated using string-overlap measures against reference translations; such measures can give an overly-rosy picture of quality \citep{kocmi-2021-ship,caswell-2025-smol}. 

We address this issue by studying the ability of LLMs to translate between sentences of \textit{formal languages} as a proxy for ICMT for natural languages. We parameterize and generate linguistically-interpretable Synchronous Context-Free Grammars (SCFGs; \citealt{aho-1969-syntax}) which each define a pair of Context-Free Languages (CFLs; \citealt{chomsky-1963-algebraic}). Using these SCFGs we sample paired sentences from the languages and construct a formal analogue to  ICMT: we provide the model with the grammar and ask it to translate between the languages.

With natural languages, it is hard to isolate which factors make grammar-based translation difficult for LLMs. Since languages may share some but not all features with one another---word order, morphology, vocabulary through common descent or loans, writing conventions, and so on---it is possible that models may ignore the provided grammatical descriptions in favor of producing token sequences that are similar to grammatical sentences from other, high-resource languages seen during training. Our formal analogue of ICMT allows us to manipulate these factors in a controlled way, something that would not be possible with natural languages.

Using this setup, we evaluate how well GPT-5's \citep{openai-2025-introducing-a} and Gemma 3's \citep{gemmaTeam-2025-gemma-a} abilities to perform ICMT are affected by properties of the setup, with the following findings:
\begin{compactenum}
    \item \textbf{Grammar Size.} While some models fare well when grammars are very small (hundreds of words), performance falls sharply as they approach the sizes needed to model human language (several thousand).
    \item \textbf{Sentence Length.} While models are capable of translating very short sentences ($<$10~words), performance drops noticeably once sentences are longer than 20~words.
    \item \textbf{Word Order.} When translating from a subject-verb-object language (SVO, like English), the word order of the target language does \emph{not} significantly impact model performance.
    \item \textbf{Morphology.} Models are significantly better at translating between two languages which do not mark person and number agreement, and are worst at translating from a language which does not mark agreement into one that does.
    \item \textbf{Orthography.} Performance drops off as the written representation of the target languages becomes less frequent: models are best at translating into languages which use the Latin script, worse at ones which use the Cyrillic script, worse still at ones which use the Hebrew script, and uniformly terrible at ones which use the Hebrew script with vowel markings.
\end{compactenum}
In all conditions, we observe that the accuracy of translations is strongly and significantly overestimated by several string-overlap heuristics when compared to the exact-match accuracy provided by having the correct translation generated by the formal grammar. Our findings suggest that while models can in fact make use of in-context descriptions of naturalistic languages to translate strings, contra \citet{aycock-2024-llms}, their robustness in this task is limited by the complexity of the grammars used to define the languages and of the sentences being translated.

\section{Background}

\paragraph{(In-Context) Machine Translation} Language models have been explored as tools for machine translation, either as an objective they are directly trained for using large parallel corpora of equivalent sentences in the source and target language \citep{kalchbrenner-2013-recurrent}, or else as a capability arising from LLMs that have been trained on unsupervised corpora of monolingual texts and later been tuned with minimal paired data to translate between languages they have become proficient in \citep{lample-2017-unsupervised}. Either setting requires a large volume of training data in each language, limiting the efficacy of language models for translation into or out of low-resource languages. This scarcity of necessary data is partially addressed by strategies like forward- \citep{zhang-2016-exploiting} and back-translation \citep{sennrich-2016-improving} to generate synthetic bitexts as a substitute for natural parallel corpora, but the efficacy of these approaches is limited by the quality of the synthetic data \citep{przystupa-2019-neural, graca-2019-generalizing}, which in turn is highly dependent on the availability of natural gold data \citep{burlot-2018-using, edunov-2018-understanding, wu-2019-extract}. More recent work has focused on leveraging increasingly-capable base models as foundations for training or prompting specialized translation systems \citep{alves-2024-tower,nllbTeam-2024-scaling,dang-2024-aya,ataman-2025-machine,omnilingualMtTeam-2026-omnilingual}. Some of these improvements transfer to the low-resource setting, especially when paired with intensive work to expand the training corpus for underrepresented languages, but large gaps in translation quality still remain between model performance in high-resource and low-resource settings \citep{caswell-2025-smol,omnilingualMtTeam-2026-omnilingual}.

\begin{figure*}
$$
\begin{Bmatrix*}[l]
\mathrm{S} \to \langle \NP\,\VP,\NP\,\VP \rangle \\
\VP \to \langle \VB\,\NP,\NP\,\VB \rangle \\
\NP \to \langle \tok{I}, \tok{watashi wa} \rangle \\
\NP \to \langle \tok{the box}, \tok{hako wo} \rangle \\
\VB \to \langle \tok{open}, \tok{akemasu} \rangle \\
\end{Bmatrix*} \quad \sim \quad
\begin{matrix*}[l]
    \textrm{EN:} & \tok{I open the box} \\
    \textrm{JP:} & \tok{watashi wa hako wo akemasu}
\end{matrix*}
$$
    \caption{(\textbf{left}) A small SCFG defining fragments of English and Japanese using production rules, from which is sampled (\textbf{right}) a corresponding pair of English and Japanese sentences. Adapted from \citet{chiang-2021-synchronous}.}
    \label{fig:demo-scfg}
\end{figure*}

A potential way to circumvent this dependence is to give models sufficient information about the low-resource language \emph{in-context} rather than in training. In this paradigm, an LLM which can make sufficient use of its context window could be prompted with a complete description of one or more languages and a parallel word list and translate sentences on this basis alone, without the need for additional training. Since such descriptions, like reference grammars, are far more compact than pretraining corpora, they are plausible as sources of linguistic knowledge for even extremely-low-resource languages. This approach to in-context machine translation was first explored by \citet{tanzer-2023-benchmark} which introduces a benchmark for Machine Translation from One Book (MTOB) for Kalamang, a language spoken on Indonesian Papua that has virtually no written corpus but which is well-described by a linguistic reference grammar \citep{visser-2022-grammar}. \citet{tanzer-2023-benchmark}, and subsequently \citet{geminiTeam-2024-gemini}, compare the performance on then-frontier LLMs on the English-Kalamang MTOB task against that of a human who is not a speaker of Kalamang but who is given access to the same materials. Both works find that LLMs demonstrate non-trivial performance at this task, but that they fail to match the (non-native-speaker) human baseline.

This result is further complicated by ablations \citep{tanzer-2023-benchmark} and subsequent investigations \citep{aycock-2024-llms} which find that parallel sentence corpora within the in-context language descriptions are critical for LLMs' success, while models gain significantly less from the grammatical descriptions themselves. It is also difficult to interpret the reported improvement in absolute terms since the difficulty and cost of obtaining native-speaker judgments of the quality of translations from or to extremely low-resource languages means that evaluations need to rely on string-overlap measures of correctness compared to human baselines from non-native speakers.

\paragraph{(Synchronous) Context-Free Grammars} A Context-Free Grammar (CFG, \citealt{chomsky-1963-algebraic}) is a generative model of a formal language defined by a vocabulary of words $\Sigma = \{\alpha, \beta, \gamma, \dotsc\}$, a set of non-terminal symbols $V = \{A, B, C, \dotsc\}$, a privileged start symbol $\tok{S} \notin V$, a set of non-terminal production rules of the form $A \to B\, C$, and a set of terminal production rules of the form $A \to \alpha$. Originally designed as a model for natural language syntax, they are also widely used to define the syntax of programming languages in Backus--Naur Form \citep{Backus1959-gq}.

Synchronous CFGs are an extension in which a single grammar defines a \emph{pair} of context-free languages based on equivalent production rules (\citealt{aho-1969-syntax}; see \citealt{chiang-2021-synchronous} for an expository treatment). Non-terminal production rules take the form of pairs $A \to \langle B\,C, D\, E \rangle$, where the left half of the tuple defines a context-free production in the first language while the right half defines an equivalent production in the second language. Terminal production rules are likewise defined as pairs $A \to \langle \alpha, \beta \rangle$ where $\alpha$ is a word in the first language and $\beta$, a word in the second. This formulation allows for the sampling of pairs of equivalent strings in each language defined by the grammar, as shown in~\cref{fig:demo-scfg}.

\section{Methodology}

We leverage SCFGs as a formal abstraction for the kinds of linguistic reference grammars of natural languages used in \citet{tanzer-2023-benchmark}; this move to a formal domain provides several benefits:
\begin{compactenum}
    \item \textbf{Easy verification.} Since SCFGs are generative models of the languages they define, every sentence $\mathbf{s}$ sampled from one of the grammar's languages comes paired with a gold translation $\mathbf{t}$. This lets us easily verify if a predicted translation $\mathbf{\hat{t}}$ matches its target exactly;
    \item \textbf{Tunable parameterization.} We can vary hyperparameters of each grammar, its defined languages, and the sentences sampled from each to determine how these factors impact performance. The hyperparameters we vary are:
    \begin{itemize}
        \item the grammar size $\abs{G}$, measured by the number of production rules;
        \item the source and target sentence lengths $\ell_s,\ell_t$, measured by the number of space-separated words in each;
        \item the basic word order of the source and target languages $W_s,W_t$, varying between subject--verb--object (SVO, akin to English), subject--object--verb (SOV, akin to Basque), and object--verb--subject (OVS, akin to Hixkaryana); see~\cref{tab:wordorder} for examples of these different orderings;
        \item whether or not person and number features are represented in verbal morphology; see~\cref{tab:agreement} for examples;
        \item the orthography of the source and target language (Latin, Latin with diacritics, Cyrillic, Hebrew, and Hebrew with vowel pointing); see~\cref{tab:scripts} for examples.
    \end{itemize}
    \item \textbf{Guarantee of no data contamination.} We generate the vocabularies for each language from scratch, creating novel naturalistic languages which are guaranteed to have never appeared in an LLM's training corpus. This ensures that a model's success on the task is not due to having been exposed to either the source or target language (or any closely-related languages) during training.
\end{compactenum}

\begin{table}[ht]
    \centering \small
    \begin{tabularx}{\linewidth}{lcX}
        \toprule
        \textbf{Language} & \textbf{Word Order} & \textbf{Example Sentence} \\ \midrule
        English & \subj{S}\verbal{V}\obj{O} & the \subj{jaguar} \verbal{ate} the \obj{man} \\
        Basque & \subj{S}\obj{O}\verbal{V} & \subj{jaguarrak} \obj{gizona} \verbal{jan} du \hfill {\small Personal communication} \\
        Hixkaryana & \obj{O}\verbal{V}\subj{S} & \obj{toto} \verbal{yonoye} \subj{kamara} \hfill {\small\citealt[p.~593]{derbyshire-1977-word}} \\
        \bottomrule
    \end{tabularx}
    \caption{Languages can differ in the ordering between \subj{subjects}, \verbal{verbs}, and \obj{objects}.}
    \label{tab:wordorder}
\end{table}

\begin{table}[ht]
    \centering
    \begin{tabularx}{\linewidth}{>{\small}l@{\hspace{4pt}} >{\small}c >{\small}c >{\small}c >{\small}c}
      \toprule
      & {\bfseries NoAgr $\to$ NoAgr}
      & {\bfseries Agr $\to$ NoAgr}
      & {\bfseries Agr $\to$ Agr}
      & {\bfseries NoAgr $\to$ Agr} \\
      \midrule
      {\fade{1sg}}
      & \pron{na} lam $\to$ \pron{ni} tor
      & \pron{na} lam\suff{mi} $\to$ \pron{ni} tor
      & \pron{na} lam\suff{mi} $\to$ \pron{ni} tor\suff{ik}
      & \pron{na} lam $\to$ \pron{ni} tor\suff{ik} \\
      {\fade{3sg}}
      & \pron{sa} lam $\to$ \pron{su} tor
      & \pron{sa} lam\suff{su} $\to$ \pron{su} tor
      & \pron{sa} lam\suff{su} $\to$ \pron{su} tor\suff{o}
      & \pron{sa} lam $\to$ \pron{su} tor\suff{o} \\
      {\fade{3pl}}
      & \pron{ran} lam $\to$ \pron{ren} tor
      & \pron{ran} lam\suff{sar} $\to$ \pron{ren} tor
      & \pron{ran} lam\suff{sar} $\to$ \pron{ren} tor\suff{on}
      & \pron{ran} lam $\to$ \pron{ren} tor\suff{on} \\
      \bottomrule
  \end{tabularx}
    \caption{Different person/number agreement paradigms between the source and target languages; verbal \suff{suffixes} in orange optionally reflect the person (1st or 3rd) and number (singular or plural) of the \pron{pronoun}.}
    \label{tab:agreement}
\end{table}

\begin{table}[ht]
    \centering
    \includegraphics[width=\linewidth]{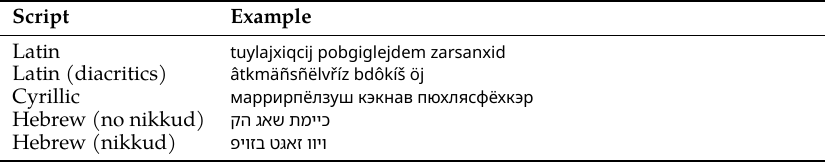}
    \caption{Examples of target sentence fragments in the various scripts.}
    \label{tab:scripts}
\end{table}

We generate SCFGs by defining a meta-grammar which parameterizes the source and target languages according to the features mentioned above. An abbreviated example of one such grammar can be found below in~\cref{fig:scfg}, and in full in~\cref{sec:prompts}. These resulting grammars group production rules into linguistically-interpretable phrases like complementizer phrases (CPs), tense phrases (TPs), noun phrases (NPs), verb phrases (VPs), and so on. This ensures that our formal grammars are similar to the kinds of structural descriptions given in natural-language reference grammars.

\begin{figure}[ht]
    \centering
    \begin{minipage}{0.45\linewidth}
\begin{lstlisting}[style=grammartt]
S -> <CP_matrix, CP_matrix>
CP_matrix -> <CNULL TP, CNULL TP>
CP_embed -> <C TP, C TP>
TP -> <NP_SUBJ TBAR, TBAR NP_SUBJ>
TBAR -> <T VP, VP T>
NP_SUBJ -> <PRON, PRON>
NP_SUBJ -> <PROPN, PROPN>
NP_SUBJ -> <DP, DP>
VP -> < VBAR, VBAR >
VBAR -> <V OBJ_PHRASE, OBJ_PHRASE V>
...
\end{lstlisting}
    \end{minipage}
    \begin{minipage}{0.45\linewidth}
\begin{lstlisting}[style=grammartt]
V -> <'yuydakvovxer', 'fehdexfum'>
V -> <'vejdetwukwesfef', 'gunlezmirtujcib'>
V -> <'dilqegzutcas', 'vuhledkobwos'>
V -> <'rofxew', 'tuvrol'>
V -> <'bisjez', 'kerpib'>
N -> <'kuqxasyuc', 'yuclajmezlosqil'>
ADJ -> <'rizwevjewkas', 'wekjil'>
ADJ -> <'rutsal', 'bodnodfuwtad'>
ADJ -> <'diylam', 'daptar'>
...
\end{lstlisting}
    \end{minipage}
    \caption{An abridged SCFG defining two naturalistic formal languages.}
    \label{fig:scfg}
\end{figure}

Given an SCFG $G$ defining a pair of source $L_s$ and target $L_t$ languages we can sample a pair of corresponding sentences $(\mathbf{s}, \mathbf{t})$, where $\mathbf{s} = [s_1, s_2, s_3, \dotsc]$ is derived from the production rules of the source language $L_s$ and $\mathbf{t}$, from those of the target language $L_t$. We then give an LLM the grammar--sentence pair $(G, \mathbf{s})$ and prompt it to translate $\mathbf{s}$ into $\mathbf{t}$ using the rules of $G$. To sample sentences of increasing length, our grammars allow sentences to have arbitrarily many nested clauses just as they do in natural languages.

We primarily compare the model's translation $\mathbf{\hat{t}}$ to the gold-production $\mathbf{t}$ produced by the grammar using exact-match accuracy as a proxy for native speaker judgments, where a model scores $1$ if $\mathbf{\hat{t}} = \mathbf{t}$, or $0$ otherwise. In~\cref{sec:detailed-results} we report model results using three additional metrics: \emph{bag-of-words accuracy}, which compares the (unordered) multiset of words in $\mathbf{\hat{t}}$ and $\mathbf{t}$ to see if models get translations right modulo word order; and two string overlap heuristics, $\BLEU$ \citep{papineni-2001-bleu} and $\chrFPP$ \citep{popovic-2017-chrf++} which are widely used in machine translation literature.

We use our formal string transduction task to evaluate a variety of LLMs, including OpenAI's GPT-5 series (\texttt{gpt-5}, \texttt{gpt-5-mini}, and \texttt{gpt-5-nano}), Google's open-weight Gemma 3 series (\texttt{gemma-3-1b-it}, \texttt{gemma-3-4b-it}, and \texttt{gemma-3-12b-it}).

\section{Results}

\begin{figure*}[ht]
    \centering
    \includegraphics[width=\linewidth]{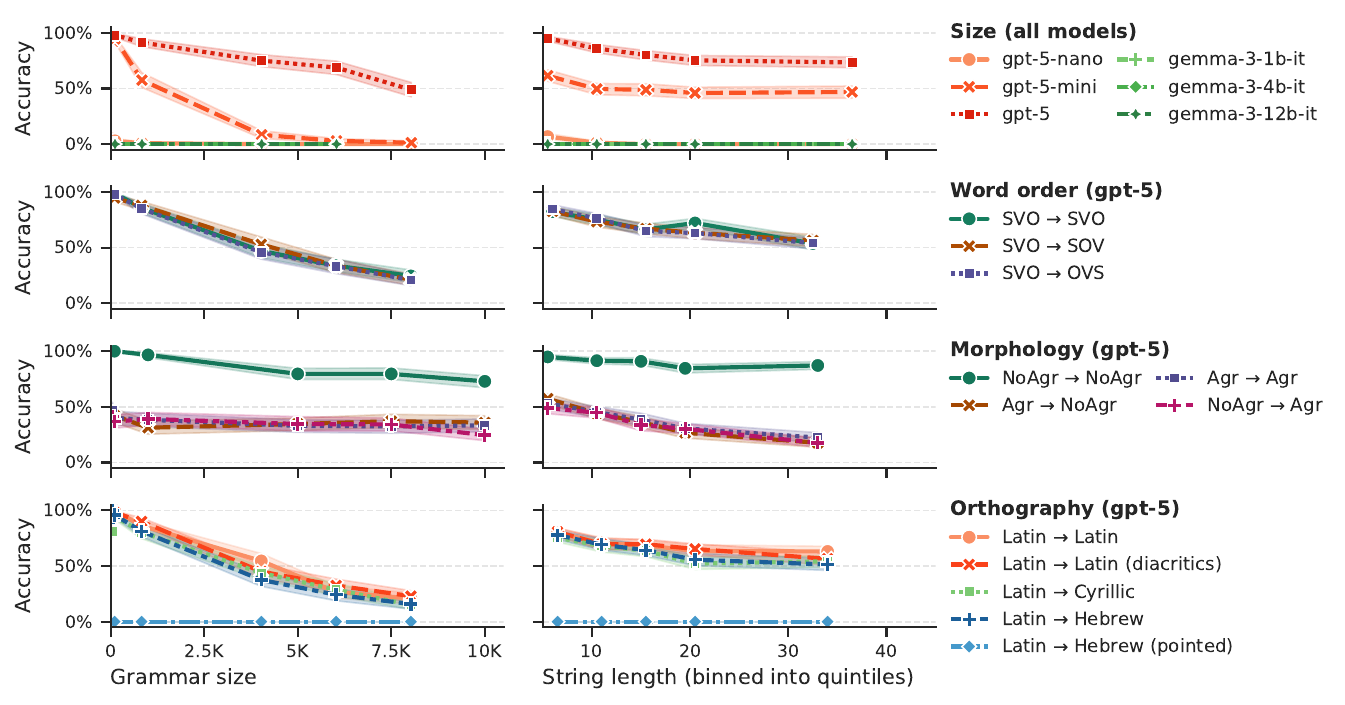}
    \caption{Mean accuracy on size (all models) and word order, morphology, and orthography (\texttt{gpt-5} only) experiments; error bars show 95\% confidence intervals. We find that grammar size and sentence length, a language's morphology type, and its orthography impact translation accuracy, while its basic word order does not.}
    \label{fig:experiments}
\end{figure*}

\subsection{Larger grammars and longer sentences make translation harder} \label{ssec:size}

We first investigate how the grammar size (defined by the number of rules in the grammar) and sentence length (defined by the number of space-separated words) impact a model's performance. We generate grammars of sizes $25 \leq \abs{G} < 10k$ and from these grammars sample sentence pairs where each sentence has length $3 \leq \ell \leq 50$. We hold all other parameters constant at the following values: both the source and target language are lexicalized using Latin-script characters; and the source language has SVO word order while the target language has SOV word order (akin to translating between English and Basque).

Models perform worse on larger grammars (\cref{fig:experiments}, row 1, left) and longer sentences (\cref{fig:experiments}, row 1, right). On the shortest sentences and smallest grammars, \texttt{gpt-5} and \texttt{gpt-5-mini} attain near-perfect accuracy. As the size of the grammar increases, model performance drops sharply. A similar, though less stark trend, is found for the impact of the input length.

\subsection{Grammatical properties of languages can affect performance} \label{ssec:structure}

Languages differ in their grammatical properties, such as their word order (e.g., do subjects precede verbs or vice-versa) or the degree to which they display inflectional morphology (e.g., do verbs conjugate to reflect the person and number of the subject). Since language models are trained on available language data, it is possible that this exposure may instill a bias for certain grammatical properties over others which could override the explicit guidance given in an in-context description of the language. This worry is compounded by the fact that these grammatical properties are not equally distributed across the world's languages, nor across the corpora used for language model (pre)training: cross-linguistic surveys show that SOV languages comprise 40--43\% of the world's languages, SVO languages comprise 35--40\%, while OVS languages comprise less than 1\% \citep{dryer-2013-order,hammarstrom-2016-linguistic}.

\paragraph{Word Order.} We investigate whether models are affected by differences in basic word order between the source and target languages. We hold the source language word order fixed at SVO and vary the target word order between SVO, SOV, and OVS. We restrict the languages to use Latin script. We find that differences in the word order between the source and target languages have negligible effects on the performance of language models across either grammar sizes (\cref{fig:experiments}, row 2, left) or sentence lengths (\cref{fig:experiments}, row 2, right), indicating that models are capable of reordering words based on the provided grammatical structure rules.

\paragraph{Morphology.} We also investigate whether morphological properties of the source and target languages affect model performance. Since languages with minimal morphology convey less information on individual words than those with rich morphology, it may be the case that language models struggle at translating between or into languages with extensive inflection. To test this, we systematically vary whether grammatical person and number are represented overtly in the morphology of verbs in the source and target languages, as shown in~\cref{tab:agreement}.

We find that the agreement paradigm strongly impacts models' translation accuracy, as shown in~\cref{fig:experiments}, row~3. Language pairs where neither language has overt agreement morphology on verbs are the easiest, while translating from a language without agreement morphology into one with morphology is the hardest. Note that for these cases (NoAgr $\to$ Agr), since the lexical mapping between the source and target languages is now one-to-many, we credit models for producing \emph{any} possible translation of the source sentence regardless of whether the features match, even though most cases can be disambiguated by looking at the person and number features of the relevant noun or pronoun.

\subsection{Models are affected by the written representation of languages}

Similar to how the preponderance of training data may bias models to prefer certain grammatical structures in translations over the specification of the target language in the grammar, the written representation of the source and target language may influence model performance; of particular concern are languages whose orthographic conventions may be different from ones present in training data. To test this, we hold the source language fixed to Latin script and vary the target language between Latin, Cyrillic, and Hebrew scripts. For Latin and Hebrew, we include versions with and without diacritical marks (known as \emph{nikkud} or \emph{pointing} in the case of Hebrew); see \cref{tab:scripts} for examples of target sentences in the different scripts. As before, we use SVO word order for both the source and target languages.

We find that orthography has a strong impact on model performance; models perform best
at translating into Latin script, moderately worse at translating into Cyrillic, and worst at translating into Hebrew. Translation into Hebrew fails completely when the target orthography includes vowel pointing, with models attaining 0\% exact-match accuracy for all sentence lengths and grammar sizes. When the vowel pointing is removed, performance improves but remains quite poor, below that of Cyrillic and Latin scripts. This suggests that models are influenced by their tendency to produce $n$-gram token sequences which are similar to those encountered in training data \citep{mccoy-2024-embers}, even when the exact information needed for translation is provided in-context.

\section{Error Analysis}

\begin{figure}[ht]
    \centering
    \includegraphics[width=\linewidth]{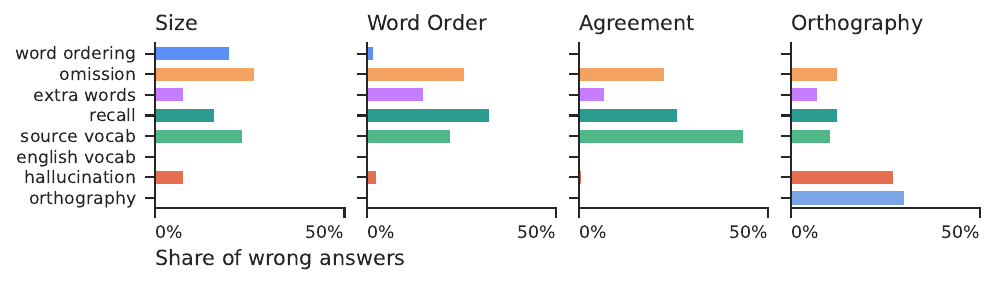}
    \caption{Error distributions for \texttt{gpt-5} by experiment. Error categories are not mutually exclusive. The most frequent error types are \emph{source vocabulary} words leakage, \emph{recall errors}, \emph{omission} of words, and in the case of the orthography experiment, \emph{hallucinating} vocabulary words and translating words in the wrong \emph{orthography}.}
    \label{fig:err-taxonomy}
\end{figure}

We categorize the errors that models make and analyze their distribution. As~\cref{fig:err-taxonomy} shows below for \texttt{gpt-5}, the majority of errors committed are \emph{recall} errors, in which the model mistranslates a term from the source language into a word from the target language vocabulary that is different from the correct translation (akin to mistranslating the English \emph{cat} as \emph{chien} `dog' instead of the correct \emph{chat} `cat' in French); \emph{source vocabulary} leakage, in which the model copies a source-language word into the translation  unchanged instead of translating it; and \emph{omission}, in which models fail to include all necessary target-language words. In the orthography experiments we find that models will commit \emph{orthographic} errors, where they produce characters or entire words in the wrong script, or \emph{hallucinate} entirely new vocabulary words which are not present in the source or target language vocabularies. Below, we include a taxonomy of the major error types used here along with examples from model outputs exemplifying those errors.

\begin{table}[ht]
    \centering
    \includegraphics[width=\linewidth]{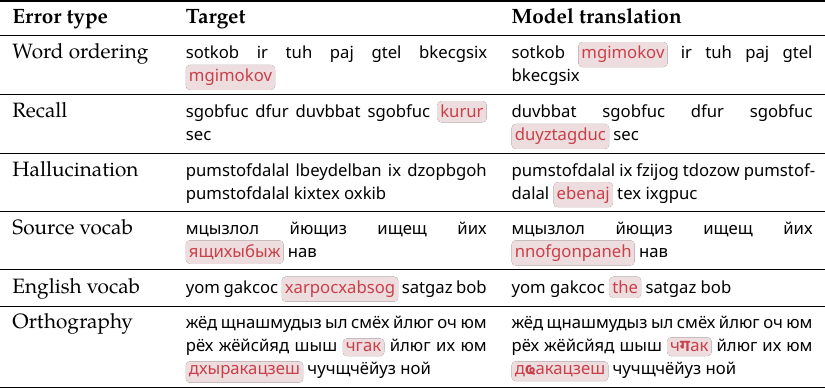}
    \caption{Examples of the various error types; words of note are highlighted in red, with specific characters (Devanagari GA \texttt{U+0917}; Abkhazian HA \texttt{U+04A9}) bolded.}
    \label{tab:error-table}
\end{table}

\paragraph{Word Order.} Models sometimes produce translations which are incorrect only because the word order of the output does not match the word order defined by the grammar. We identify these errors by finding incorrect translations whose (unordered) set representation is equal to that of the target sentence.

\paragraph{Recall Error.} Models sometimes mistranslate individual words in the source language, leading to translations which are the right length but contain the wrong words. We identify these errors by finding incorrect translations which contain target-language words present in the target-language vocabulary but not in the specific target sentence.

\paragraph{Hallucination.} Models sometimes hallucinate vocabulary words in the target language which are not present in the grammar. We identify these by finding incorrect translations which contain words present in neither the source nor the target language vocabularies. These errors can sometimes take the form of \emph{misspellings}, where a model produces an almost-correct word except for a few characters; or they can invent new words wholesale.

\paragraph{Source Vocabulary.} Models sometimes fail to translate individual words from the source language, leaving them in the predicted translation for the target language. We identify these by finding translations which contain terms present in the source language vocabulary but not in the target language vocabulary.

\paragraph{Orthography Errors.} Models sometimes translate into the wrong orthography, producing characters or whole words which use symbols from another script. We identify these by finding translations which contain Unicode codepoints outside the expected range for the target language, or which additionally \emph{fail} to contain any codepoints from the diacritic ranges (in the case the Latin and Hebrew variants with diacritic marks).

\paragraph{English Vocabulary.} Models occasionally include English words in their translations instead of words from the source language.

\section{Discussion}

Our results suggest that current LLMs can use explicit formal descriptions of a language to perform translation, but that performance is influenced by a number of features of the languages and grammars. Tempering to the findings of \citet{aycock-2024-llms} that LLMs could not use in-context language descriptions without parallel sentences for translation, the positive results for current models, especially on shorter sentences and smaller grammars, demonstrate that LLMs \emph{can} make use of such resources for translation.

The models' performance is tempered most strongly by grammar size and sentence length. By contrast, differences in source and target word order do not meaningfully reduce performance. This suggests that the dominant bottleneck is not the need to learn a particular cross-linguistic mapping such as SVO-to-SOV reordering. Rather, the harder problem is maintaining and correctly applying a larger inventory of symbolic rules over longer derivations.

The orthography results point to a second, distinct bottleneck. Performance falls sharply when the target language is written in a script that differs from the source, and it collapses entirely for the least frequent script, fully pointed Hebrew. Because the underlying grammars and derivations are held constant across these conditions, this degradation cannot be explained by syntactic difficulty alone. The likely implication is that ICMT depends not just on abstract rule induction, but also on a model's robustness in copying, segmenting, and emitting unfamiliar character sequences. For practical ICMT, this means that even if a model can infer the right structural mapping from a grammar, orthographic unfamiliarity may still prevent successful translation. Errors of this kind also likely indicate brittleness in LLMs' underlying capabilities, where success depends on the sequence of output tokens having a high degree of $n$-gram similarity to token sequences encountered in training, irrespective of the task definition \citep{mccoy-2024-embers}.

\subsection{Limitations}

Our SCFG formalization deliberately abstracts away from many aspects of natural-language translation. It does not capture ambiguity at the level typically found in natural-language descriptions, many types of phrase structure found in natural language (e.g., our grammars do not contain prepositional phrases), or interactions between syntax and semantics that make many translation decisions under-determined without discourse context. Our grammars also enforce unusually direct lexical correspondences between the two languages. Even when lexical items may span multiple orthographic words, each source-side terminal is paired with a specific target-side terminal. Natural-language translation often involves lexical gaps, many-to-one paraphrases, and semantically-conditioned alternations that are not well approximated here. The present results should therefore be interpreted as an upper bound on one component of ICMT, namely the ability to execute relatively clean symbolic transductions from explicit grammatical rules.

Finally, we deliberately study a setting in which the model receives only a grammar and an input sentence. This isolates whether models can use formal descriptions directly, but omits example translations or miniature parallel corpora. As a result, our experiments do not address how grammatical descriptions and few-shot exemplars interact, nor whether examples can compensate for the weaknesses we observe with larger grammars or unfamiliar scripts.

\bibliographystyle{colm2026_conference}
\bibliography{paperpile}

\appendix
\section{Example Prompts} \label{sec:prompts}

\begin{prompt}
You will be presented with a synchronous context-free grammar (SCFG) which defines a mapping between two context-free languages.You will also be presented with a sentence produced by one of the languages defined by the grammar. You task is to use the rules of the grammar to translate the sentence from the source language into the target language.

A grammar is defined by a set of production rules. Rules come in two forms: non-lexical rules, of the form \verb|`A -> <B C, D E>`| where all of \texttt{`A, B, C, D, E`} are non-terminal symbols; and lexical rules, of the form \verb|`A -> <'a', 'b'>`|  where \verb|`A`| is a non-terminal symbol and \verb|`'a'`| and \verb|`'b'`| are terminal symbols (words). The right-hand side of each production rule consists of a pair demarcated by angle brackets. The first element of this pair shows the expansion of the left-hand side in one language, and the second element shows the expansion in the other language. The order of the symbols may differ between the two languages. All grammars are guaranteed to start with a distinguished start symbol \verb|`S`|. All grammars are defined according to X-bar style rules, intended to model natural language syntax. This means that productions are built are phrases (XP) which produce specifiers (YP) and bar-level projections (XBar); these bar-level projections in turn produce heads (X) and complements (ZP). Certain lexical productions in the grammar produce words which begin with a null symbol \verb|'\u2205'|; these words are phonetically null and do not appear in the surface forms of either the input or output sentences, though they may be important for the syntactic structure of the sentence. Do not include these null words in your output sentence, though you may need to reason about them to get the correct structure.

You may use any reasoning strategy you like to solve this task, including identifying the categories of the words in the input sentence, using the grammar to build a parse tree for the input, and then following that derivation using the other language's expansions to produce the output sentence. Feel free to write down intermediate steps in your reasoning.

You will be evaluated based on the string accuracy of the output sentence, which you should format like the following: \verb|`Final answer: <output sentence>`|. If you do not end your response with this format, you will be marked as incorrect.

Here is the synchronous context-free grammar:

\begin{verbatim}
```
S -> <CP_matrix, CP_matrix>
CP_matrix -> <CNULL TP, CNULL TP>
CP_embed -> <C TP, C TP>
TP -> <NP_SUBJ TBAR, TBAR NP_SUBJ>
TBAR -> <T VP, VP T>
NP_SUBJ -> <PRON, PRON>
NP_SUBJ -> <PROPN, PROPN>
NP_SUBJ -> <DP, DP>
VP -> < VBAR, VBAR >
VBAR -> <V OBJ_PHRASE, OBJ_PHRASE V>
OBJ_PHRASE -> <DP, DP>
OBJ_PHRASE -> <CP_embed, CP_embed>
DETP -> < DETBAR, DETBAR >
DETBAR -> <DET NP, NP DET>
DP -> <DP_def, DP_def>
DP -> <DP_indef, DP_indef>
DP_def -> <DET_def NP, DET_def NP>
DP_indef -> <DET_indef NP, DET_indef NP>
DP_def -> <PROPN, PROPN>
NP -> <N_HEAD, N_HEAD>
NP -> <AdjP NP, AdjP NP>
NP_COMMON -> <N, N>
NP_COMMON -> <AdjP NP_COMMON, AdjP NP_COMMON>
AdjP -> <ADJ, ADJ>
N_HEAD -> <N, N>
N_HEAD -> <PROPN, PROPN>
DET_def -> <'hayxutlutbur', 'marjiy'>
DET_def -> <'napvijnaypigtoyqix',
  'yisrazzihqomjubdibdez'>
DET_indef -> <'rucwey', 'focqersorken'>
DET_indef -> <'yosvuwnuspitqug', 'qehqojfer'>
T -> <'\u2205_T_pres', '\u2205_T_pres'>
ASP -> <'\u2205_Asp_prog', '\u2205_Asp_prog'>
V -> <'yuydakvovxer', 'fehdexfum'>
V -> <'vejdetwukwesfef', 'gunlezmirtujcib'>
V -> <'dilqegzutcas', 'vuhledkobwos'>
V -> <'rofxew', 'tuvrol'>
V -> <'bisjez', 'kerpib'>
N -> <'kuqxasyuc', 'yuclajmezlosqil'>
N -> <'voysabfuc', 'walfultev'>
N -> <'vizbuz', 'darwozliz'>
N -> <'xazcatyut', 'fubcuzpejluf'>
N -> <'potyaqpenbed', 'ligyun'>
PROPN -> <'livhuj', 'vacfaq'>
PROPN -> <'lufpar', 'hurbahcodsom'>
PROPN -> <'mebguqtar', 'jechekmuljuzhejpig'>
PROPN -> <'fedtosyoh', 'tefziw'>
PROPN -> <'sirlob', 'zatpuj'>
PRON -> <'nitxilwictof', 'piznedhufmup'>
PRON -> <'jocrixxil', 'moclabtep'>
ADJ -> <'rizwevjewkas', 'wekjil'>
ADJ -> <'rutsal', 'bodnodfuwtad'>
ADJ -> <'diylam', 'daptar'>
ADJ -> <'pinnay', 'habgaxwixsik'>
ADJ -> <'zepjuykasrom', 'wuzfej'>
C -> <'qiyben', 'goxpubpimvet'>
C -> <'fubveq', 'kajvasfelkek'>
CNULL -> <'\u2205', '\u2205'>
```
\end{verbatim}

Here is the input sentence: \verb|`sirlob rofxew livhuj`|.

Remember to end your response with the format \verb|`Final answer: <output sentence>`|.
\end{prompt} 

\section{Detailed Results} \label{sec:detailed-results}

Here we provide full results for performance on the SCFG translation task, broken down by experiment and model. We report the model's performance as measured by four metrics:
\begin{itemize}
    \item \textbf{Exact Sequence Match:} A model scores $1$ if $\mathbf{\hat{t}} = \mathbf{t}$, or $0$ otherwise.
    \item \textbf{Bag-of-Words Match:} A model scores $1$ if the (unordered) multiset of vocabulary words in $\{\hat{t}_1, \hat{t}_2, \hat{t}_3, \dotsc\}$ equals that of the gold production $\{t_1, t_2, t_3, \dotsc\}$. This removes penalties for getting the word order of a translation wrong, but still penalizes substitutions or deletions.
    \item \textbf{BLEU \textnormal{\citep{papineni-2001-bleu}}:} BLEU scores work at the word-level, catching word misordering and extra or omitted content, but is insensitive to morphological variation expressed at the character level.
    $$
    \BLEU(\hat{\mathbf{t}}, \mathbf{t}) = \BP\cdot \exp\left(\sum_{n=1}^N w_n \ln p_n(\hat{\mathbf{t}}, \mathbf{t})\right),
    $$
    where $\BP$ is a brevity penalty and $p_n$ are $n$-gram precisions weighted by $w_n$ for $n \leq N$. 
    Following standard practice, we use the parameters and implementation from SacreBLEU \citep{Post2018-sm}. 
    \item \textbf{$\chrFPP$ \textnormal{\citep{popovic-2017-chrf++}}:}
    $$
    \chrFPP(\mathbf{\hat{t}}, \mathbf{t}) = (1+\beta^2)\frac{\chrP(\mathbf{\hat{t}}, \mathbf{t})\cdot \chrR(\mathbf{\hat{t}}, \mathbf{t})}{\beta^2\cdot \chrP(\mathbf{\hat{t}}, \mathbf{t}) + \chrR(\mathbf{\hat{t}}, \mathbf{t})},
    $$
    where $\chrP$ and $\chrR$ are the word-level precision and recall, respectively, and $\beta$ is a weighting parameter that rewards recall over precision.  
    $\chrFPP$ is a character-level F-score, making it is more sensitive to intra-word morphology and less penalizing for spelling errors. Following the recommendation of \citet{popovic-2017-chrf++}, we set $\beta=2$.
\end{itemize}
These metrics each yield scores in the range $[0.0, 1.0]$ and directionally agree with one another (i.e., a score of $1.0$ is best, while $0.0$ is worst). In general, we note that the two heuristic measures ($\BLEU$ and $\chrFPP$) consistently over-estimate models' exact-match accuracies, which is to be expected, though it is perhaps worth noting that the size of this disparity frequently increases with grammar size, string length, and the relevant grammatical properties which impact model success, meaning that these heuristic measures become \emph{less accurate} predictors of model accuracy as the task becomes harder.

\subsection{Size Experiment}

\begin{figure*}[ht]
    \includegraphics[width=\linewidth]{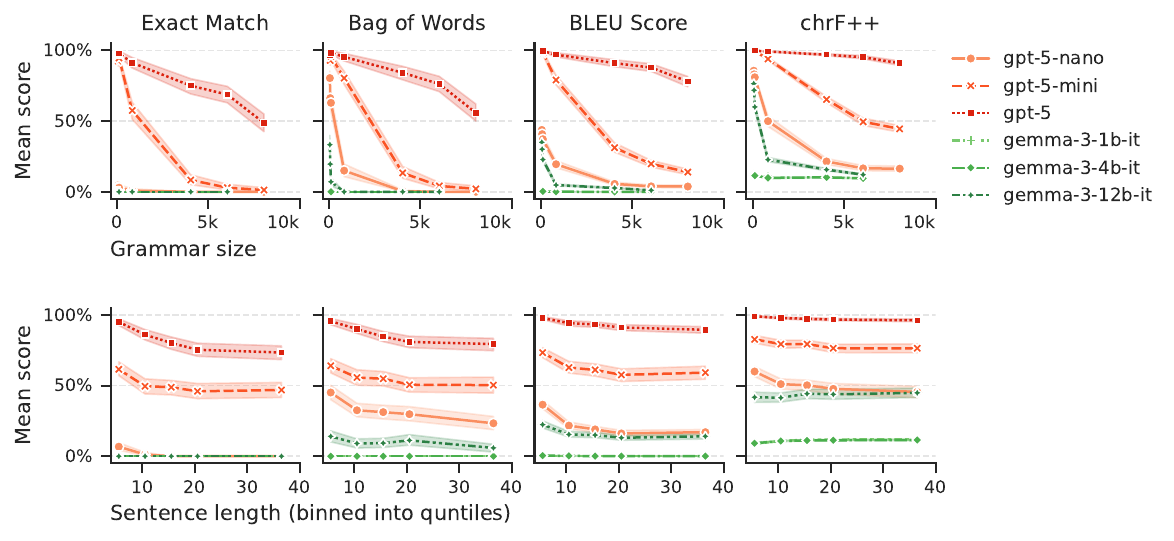}
    \caption{Full results for grammar size \& sentence length experiment. Error bars show 95\% confidence interval.}
    \label{fig:size-full}
\end{figure*}

\Cref{fig:size-full,tab:results-size,tab:results-size-length} shows results for all models on the grammar size \& sentence length experiment.

\begin{table*}[ht]
  \centering
  \small
  \sisetup{table-format=1.3}
  \begin{tabularx}{\textwidth}{>{\raggedright\arraybackslash}X l S S S S S S S}
    \toprule
    \textbf{Model} & \textbf{Metric} & \textbf{57} & \textbf{77} & \textbf{117} & \textbf{837} & \textbf{4,037} & \textbf{6,037} & \textbf{8,037} \\
    \midrule
    \multirow{4}{=}{\texttt{gpt-5}} & Exact Match & 0.967 & 0.967 & 0.979 & 0.912 & 0.750 & 0.688 & 0.487 \\
     & Bag of Words & 0.967 & 0.967 & 0.983 & 0.954 & 0.842 & 0.762 & 0.558 \\
     & $\BLEU$ & 0.994 & 0.995 & 0.995 & 0.970 & 0.910 & 0.878 & 0.780 \\
     & $\chrFPP$ & 0.998 & 0.998 & 0.998 & 0.992 & 0.970 & 0.952 & 0.909 \\
    \midrule
    \multirow{4}{=}{\texttt{gpt-5-mini}} & Exact Match & 0.975 & 0.925 & 0.946 & 0.575 & 0.083 & 0.029 & 0.013 \\
     & Bag of Words & 0.975 & 0.929 & 0.958 & 0.804 & 0.133 & 0.042 & 0.021 \\
     & $\BLEU$ & 0.994 & 0.984 & 0.984 & 0.791 & 0.312 & 0.197 & 0.139 \\
     & $\chrFPP$ & 0.997 & 0.994 & 0.994 & 0.940 & 0.653 & 0.495 & 0.443 \\
    \midrule
    \multirow{4}{=}{\texttt{gpt-5-nano}} & Exact Match & 0.046 & 0.033 & 0.029 & 0.008 & 0.000 & 0.000 & 0.000 \\
     & Bag of Words & 0.804 & 0.662 & 0.629 & 0.146 & 0.004 & 0.000 & 0.000 \\
     & $\BLEU$ & 0.438 & 0.408 & 0.372 & 0.190 & 0.055 & 0.037 & 0.036 \\
     & $\chrFPP$ & 0.856 & 0.831 & 0.810 & 0.487 & 0.209 & 0.159 & 0.158 \\
    \midrule
    \multirow{4}{=}{\texttt{gemma-3-12b-it}} & Exact Match & 0.000 & 0.000 & 0.000 & 0.000 & 0.000 & 0.000 & \multicolumn{1}{c}{---} \\
     & Bag of Words & 0.333 & 0.196 & 0.071 & 0.000 & 0.000 & 0.000 & \multicolumn{1}{c}{---} \\
     & $\BLEU$ & 0.350 & 0.301 & 0.228 & 0.048 & 0.027 & 0.010 & \multicolumn{1}{c}{---} \\
     & $\chrFPP$ & 0.765 & 0.718 & 0.600 & 0.227 & 0.157 & 0.120 & \multicolumn{1}{c}{---} \\
    \midrule
    \multirow{4}{=}{\texttt{gemma-3-4b-it}} & Exact Match & 0.000 & 0.000 & 0.000 & 0.000 & 0.000 & 0.000 & \multicolumn{1}{c}{---} \\
     & Bag of Words & 0.000 & 0.000 & 0.000 & 0.000 & 0.000 & 0.000 & \multicolumn{1}{c}{---} \\
     & $\BLEU$ & 0.003 & 0.004 & 0.003 & 0.000 & 0.000 & 0.000 & \multicolumn{1}{c}{---} \\
     & $\chrFPP$ & 0.112 & 0.122 & 0.115 & 0.099 & 0.102 & 0.093 & \multicolumn{1}{c}{---} \\
    \midrule
    \multirow{4}{=}{\texttt{gemma-3-1b-it}} & Exact Match & 0.000 & 0.000 & 0.000 & 0.000 & \multicolumn{1}{c}{---} & \multicolumn{1}{c}{---} & \multicolumn{1}{c}{---} \\
     & Bag of Words & 0.000 & 0.000 & 0.000 & 0.000 & \multicolumn{1}{c}{---} & \multicolumn{1}{c}{---} & \multicolumn{1}{c}{---} \\
     & $\BLEU$ & 0.000 & 0.000 & 0.000 & 0.000 & \multicolumn{1}{c}{---} & \multicolumn{1}{c}{---} & \multicolumn{1}{c}{---} \\
     & $\chrFPP$ & 0.108 & 0.110 & 0.111 & 0.112 & \multicolumn{1}{c}{---} & \multicolumn{1}{c}{---} & \multicolumn{1}{c}{---} \\
    \bottomrule
  \end{tabularx}
  \caption{Mean results by grammar size for all models in the size experiment.}
  \label{tab:results-size}
\end{table*}
\begin{table*}[ht]
  \centering
  \small
  \sisetup{table-format=1.3}
  \begin{tabularx}{\textwidth}{>{\raggedright\arraybackslash}X l S S S S S}
    \toprule
    \textbf{Model} & \textbf{Metric} & \textbf{5.5} & \textbf{10.5} & \textbf{15.5} & \textbf{20.5} & \textbf{36.5} \\
    \midrule
    \multirow{4}{=}{\texttt{gpt-5}} & Exact Match & 0.950 & 0.859 & 0.801 & 0.753 & 0.734 \\
     & Bag of Words & 0.955 & 0.899 & 0.846 & 0.808 & 0.793 \\
     & $\BLEU$ & 0.977 & 0.941 & 0.932 & 0.910 & 0.894 \\
     & $\chrFPP$ & 0.990 & 0.978 & 0.972 & 0.967 & 0.962 \\
    \midrule
    \multirow{4}{=}{\texttt{gpt-5-mini}} & Exact Match & 0.615 & 0.495 & 0.488 & 0.459 & 0.467 \\
     & Bag of Words & 0.640 & 0.557 & 0.548 & 0.506 & 0.502 \\
     & $\BLEU$ & 0.733 & 0.627 & 0.610 & 0.576 & 0.589 \\
     & $\chrFPP$ & 0.827 & 0.793 & 0.793 & 0.764 & 0.761 \\
    \midrule
    \multirow{4}{=}{\texttt{gpt-5-nano}} & Exact Match & 0.067 & 0.012 & 0.000 & 0.000 & 0.000 \\
     & Bag of Words & 0.447 & 0.321 & 0.307 & 0.291 & 0.226 \\
     & $\BLEU$ & 0.362 & 0.214 & 0.187 & 0.158 & 0.165 \\
     & $\chrFPP$ & 0.594 & 0.504 & 0.494 & 0.466 & 0.441 \\
    \midrule
    \multirow{4}{=}{\texttt{gemma-3-12b-it}} & Exact Match & 0.000 & 0.000 & 0.000 & 0.000 & 0.000 \\
     & Bag of Words & 0.141 & 0.090 & 0.093 & 0.112 & 0.059 \\
     & $\BLEU$ & 0.224 & 0.154 & 0.150 & 0.130 & 0.141 \\
     & $\chrFPP$ & 0.417 & 0.412 & 0.442 & 0.436 & 0.448 \\
    \midrule
    \multirow{4}{=}{\texttt{gemma-3-4b-it}} & Exact Match & 0.000 & 0.000 & 0.000 & 0.000 & 0.000 \\
     & Bag of Words & 0.000 & 0.000 & 0.000 & 0.000 & 0.000 \\
     & $\BLEU$ & 0.004 & 0.003 & 0.000 & 0.001 & 0.000 \\
     & $\chrFPP$ & 0.092 & 0.107 & 0.111 & 0.113 & 0.114 \\
    \midrule
    \multirow{4}{=}{\texttt{gemma-3-1b-it}} & Exact Match & 0.000 & 0.000 & 0.000 & 0.000 & 0.000 \\
     & Bag of Words & 0.000 & 0.000 & 0.000 & 0.000 & 0.000 \\
     & $\BLEU$ & 0.000 & 0.000 & 0.000 & 0.000 & 0.000 \\
     & $\chrFPP$ & 0.089 & 0.108 & 0.117 & 0.118 & 0.120 \\
    \bottomrule
  \end{tabularx}
  \caption{Mean results by input string length for all models in the size experiment.}
  \label{tab:results-size-length}
\end{table*}

\subsection{Word Order Experiment}

\begin{figure}[ht]
    \centering
    \includegraphics[width=\linewidth]{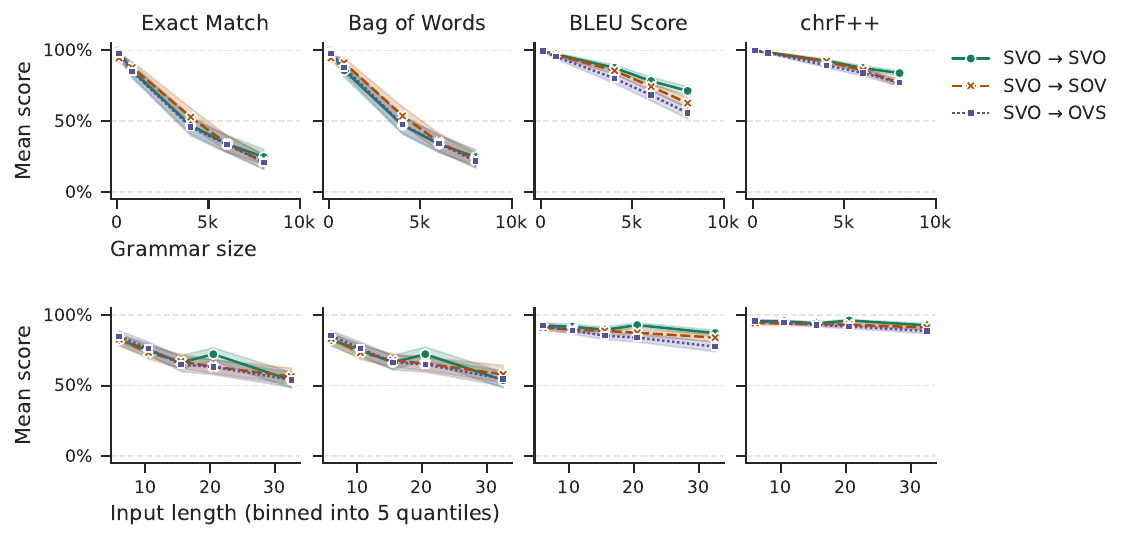}
    \caption{Full results for \texttt{gpt-5} on the word order experiment. Error bars show 95\% confidence interval.}
    \label{fig:wordorder-gpt-5}
\end{figure}

\begin{figure}[ht]
    \centering
    \includegraphics[width=\linewidth]{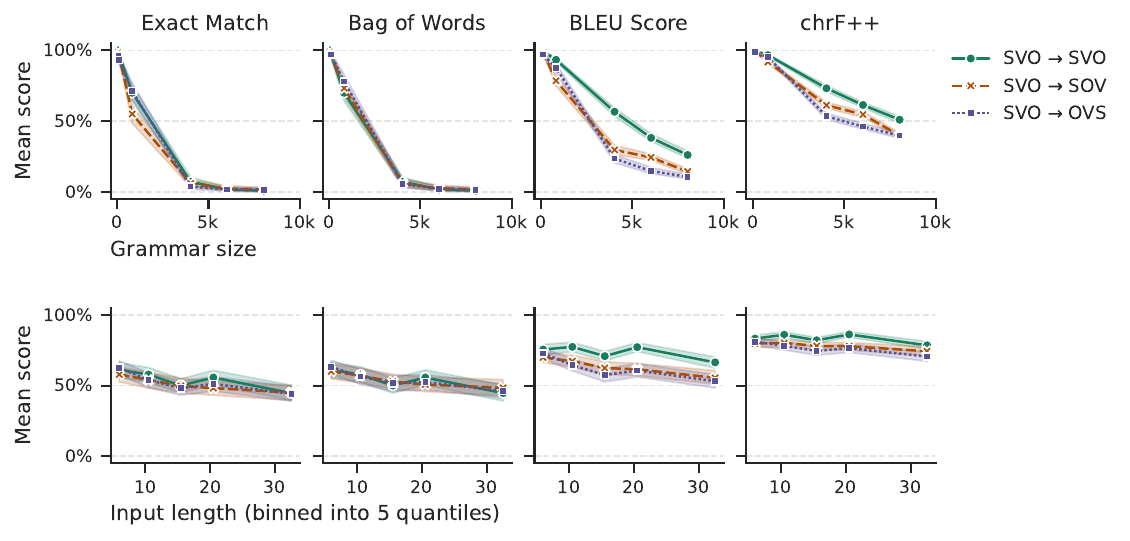}
    \caption{Full results for \texttt{gpt-5-mini} on the word order experiment. Error bars show 95\% confidence interval.}
    \label{fig:wordorder-gpt-5-mini}
\end{figure}

\begin{figure}[ht]
    \centering
    \includegraphics[width=\linewidth]{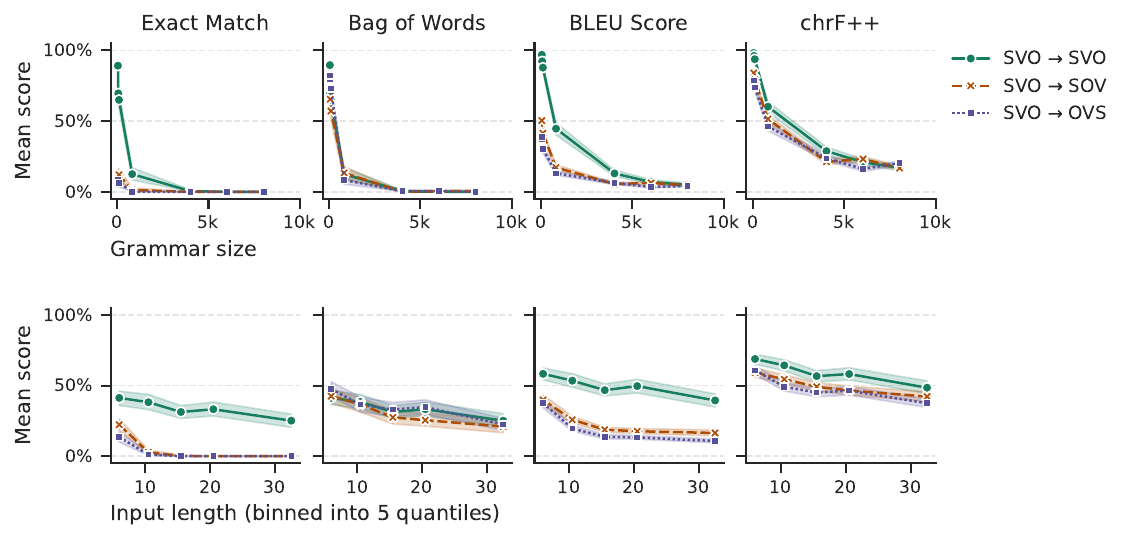}
    \caption{Full results for \texttt{gpt-5-nano} on the word order experiment. Error bars show 95\% confidence interval.}
    \label{fig:wordorder-gpt-5-nano}
\end{figure}

\begin{figure}[ht]
    \centering
    \includegraphics[width=\linewidth]{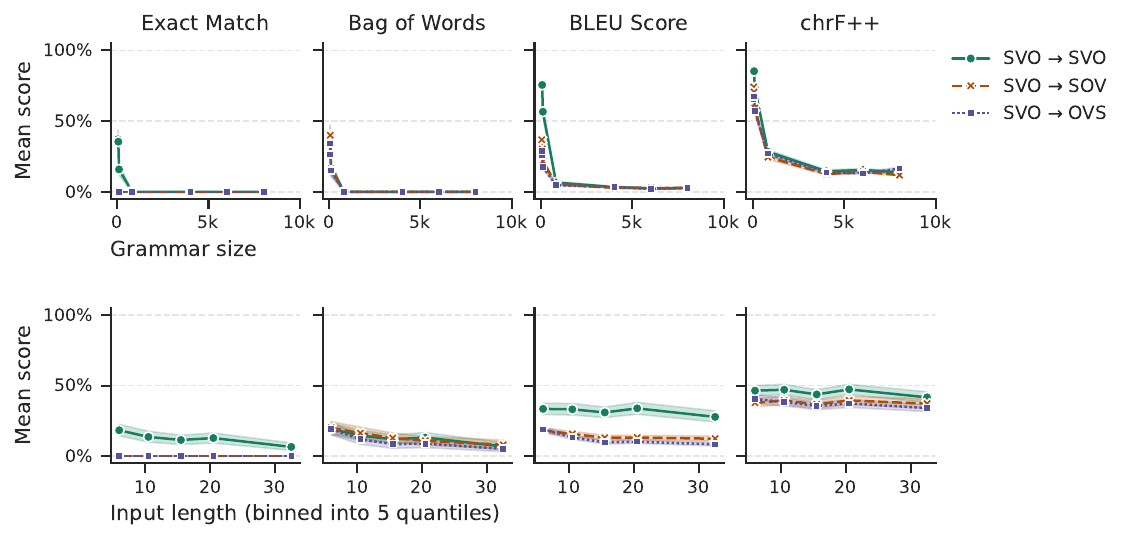}
    \caption{Full results for \texttt{gemma-3-12b-it} on the word order experiment. Error bars show 95\% confidence interval.}
    \label{fig:wordorder-gemma-3-12b-it}
\end{figure}

\begin{figure}[ht]
    \centering
    \includegraphics[width=\linewidth]{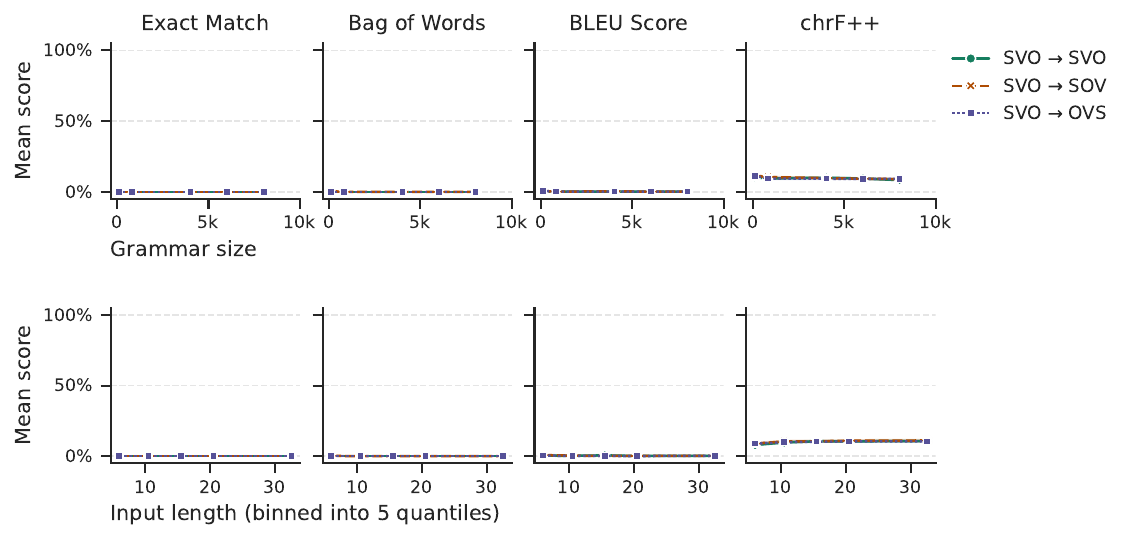}
    \caption{Full results for \texttt{gemma-3-4b-it} on the word order experiment. Error bars show 95\% confidence interval.}
    \label{fig:wordorder-gemma-3-4b-it}
\end{figure}

\begin{figure}[ht]
    \centering
    \includegraphics[width=\linewidth]{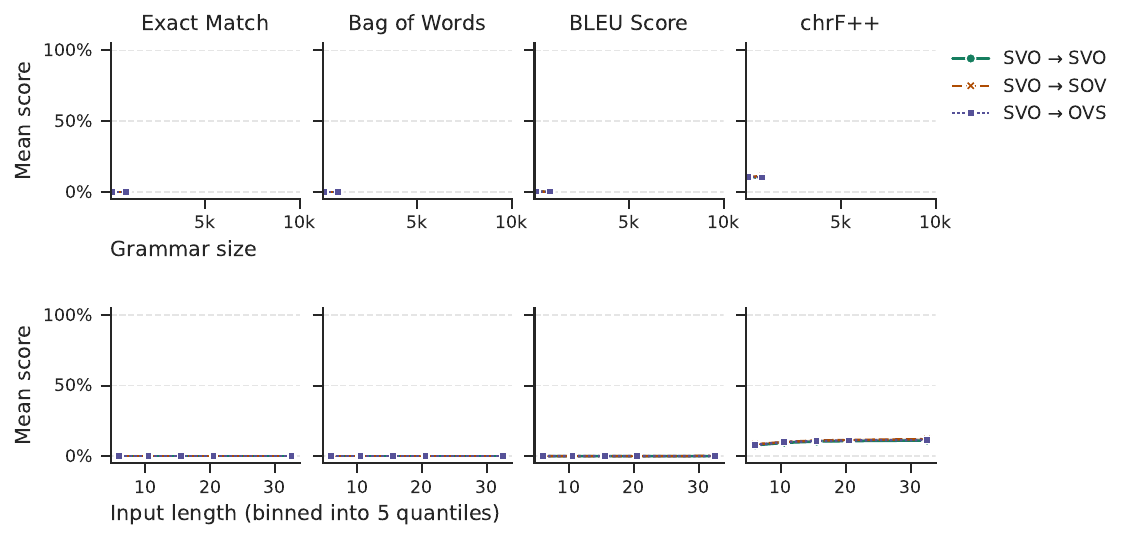}
    \caption{Full results for \texttt{gemma-3-1b-it} on the word order experiment. Error bars show 95\% confidence interval.}
    \label{fig:wordorder-gemma-3-1b-it}
\end{figure}

The figures and tables below show results on the word order experiment for \texttt{gpt-5} (\cref{fig:wordorder-gpt-5,tab:results-wordorder-grammar:gpt-5,tab:results-wordorder-length:gpt-5}), \texttt{gpt-5-mini} (\cref{fig:wordorder-gpt-5-mini,tab:results-wordorder-grammar:gpt-5-mini,tab:results-wordorder-length:gpt-5-mini}), \texttt{gpt-5-nano} (\cref{fig:wordorder-gpt-5-nano,tab:results-wordorder-grammar:gpt-5-nano,tab:results-wordorder-length:gpt-5-nano}), \texttt{gemma-3-12b-it} (\cref{fig:wordorder-gemma-3-12b-it,tab:results-wordorder-grammar:google_gemma-3-12b-it,tab:results-wordorder-length:google_gemma-3-12b-it}), \texttt{gemma-3-4b-it} (\cref{fig:wordorder-gemma-3-4b-it,tab:results-wordorder-grammar:google_gemma-3-4b-it,tab:results-wordorder-length:google_gemma-3-4b-it}), and \texttt{gemma-3-1b-it} (\cref{fig:wordorder-gemma-3-1b-it,tab:results-wordorder-grammar:google_gemma-3-1b-it,tab:results-wordorder-length:google_gemma-3-1b-it}). Note that \texttt{gemma-3-1b-it} has a much shorter maximum context window which precludes us from evaluating it on larger grammars.

\begin{table*}[ht]
  \centering
  \small
  \sisetup{table-format=1.3}
  \begin{tabularx}{\textwidth}{>{\raggedright\arraybackslash}X l S S S S S S S}
    \toprule
    \textbf{Condition} & \textbf{Metric} & \textbf{46} & \textbf{66} & \textbf{106} & \textbf{826} & \textbf{4,026} & \textbf{6,026} & \textbf{8,026} \\
    \midrule
    \multirow{4}{=}{SVO → SVO} & Exact Match & 0.996 & 1.000 & 0.988 & 0.858 & 0.471 & 0.338 & 0.246 \\
     & Bag of Words & 0.996 & 1.000 & 0.988 & 0.858 & 0.471 & 0.338 & 0.246 \\
     & $\BLEU$ & 0.999 & 1.000 & 0.997 & 0.972 & 0.877 & 0.783 & 0.714 \\
     & $\chrFPP$ & 1.000 & 1.000 & 0.999 & 0.986 & 0.925 & 0.875 & 0.839 \\
    \midrule
    \multirow{4}{=}{SVO → SOV} & Exact Match & 0.963 & 0.971 & 0.946 & 0.879 & 0.529 & 0.338 & 0.212 \\
     & Bag of Words & 0.967 & 0.971 & 0.946 & 0.912 & 0.537 & 0.350 & 0.225 \\
     & $\BLEU$ & 0.993 & 0.996 & 0.993 & 0.968 & 0.857 & 0.744 & 0.627 \\
     & $\chrFPP$ & 0.997 & 0.999 & 0.996 & 0.985 & 0.923 & 0.859 & 0.770 \\
    \midrule
    \multirow{4}{=}{SVO → OVS} & Exact Match & 0.983 & 0.988 & 0.979 & 0.850 & 0.458 & 0.333 & 0.208 \\
     & Bag of Words & 0.983 & 0.988 & 0.979 & 0.879 & 0.475 & 0.342 & 0.217 \\
     & $\BLEU$ & 0.995 & 0.999 & 0.996 & 0.956 & 0.802 & 0.683 & 0.557 \\
     & $\chrFPP$ & 0.999 & 0.999 & 0.998 & 0.983 & 0.895 & 0.841 & 0.771 \\
    \bottomrule
  \end{tabularx}
  \caption{Mean results for \texttt{gpt-5} in the word order experiment, grouped by target word-order condition and grammar size.}
  \label{tab:results-wordorder-grammar:gpt-5}
\end{table*}

\begin{table*}[ht]
  \centering
  \small
  \sisetup{table-format=1.3}
  \begin{tabularx}{\textwidth}{>{\raggedright\arraybackslash}X l S S S S S S S}
    \toprule
    \textbf{Condition} & \textbf{Metric} & \textbf{46} & \textbf{66} & \textbf{106} & \textbf{826} & \textbf{4,026} & \textbf{6,026} & \textbf{8,026} \\
    \midrule
    \multirow{4}{=}{SVO → SVO} & Exact Match & 1.000 & 1.000 & 0.971 & 0.700 & 0.071 & 0.017 & 0.008 \\
     & Bag of Words & 1.000 & 1.000 & 0.971 & 0.700 & 0.071 & 0.017 & 0.008 \\
     & $\BLEU$ & 1.000 & 1.000 & 0.996 & 0.933 & 0.566 & 0.382 & 0.261 \\
     & $\chrFPP$ & 1.000 & 1.000 & 0.998 & 0.966 & 0.732 & 0.614 & 0.510 \\
    \midrule
    \multirow{4}{=}{SVO → SOV} & Exact Match & 0.988 & 0.983 & 0.967 & 0.550 & 0.058 & 0.021 & 0.017 \\
     & Bag of Words & 0.988 & 0.988 & 0.971 & 0.733 & 0.058 & 0.025 & 0.017 \\
     & $\BLEU$ & 0.998 & 0.996 & 0.990 & 0.785 & 0.296 & 0.243 & 0.143 \\
     & $\chrFPP$ & 0.999 & 0.999 & 0.996 & 0.919 & 0.612 & 0.547 & 0.398 \\
    \midrule
    \multirow{4}{=}{SVO → OVS} & Exact Match & 0.963 & 0.963 & 0.933 & 0.713 & 0.037 & 0.017 & 0.013 \\
     & Bag of Words & 0.967 & 0.971 & 0.971 & 0.779 & 0.054 & 0.021 & 0.013 \\
     & $\BLEU$ & 0.994 & 0.987 & 0.970 & 0.877 & 0.235 & 0.147 & 0.105 \\
     & $\chrFPP$ & 0.998 & 0.994 & 0.989 & 0.952 & 0.534 & 0.461 & 0.397 \\
    \bottomrule
  \end{tabularx}
  \caption{Mean results for \texttt{gpt-5-mini} in the word order experiment, grouped by target word-order condition and grammar size.}
  \label{tab:results-wordorder-grammar:gpt-5-mini}
\end{table*}

\begin{table*}[ht]
  \centering
  \small
  \sisetup{table-format=1.3}
  \begin{tabularx}{\textwidth}{>{\raggedright\arraybackslash}X l S S S S S S S}
    \toprule
    \textbf{Condition} & \textbf{Metric} & \textbf{46} & \textbf{66} & \textbf{106} & \textbf{826} & \textbf{4,026} & \textbf{6,026} & \textbf{8,026} \\
    \midrule
    \multirow{4}{=}{SVO → SVO} & Exact Match & 0.892 & 0.696 & 0.650 & 0.125 & 0.004 & 0.000 & 0.000 \\
     & Bag of Words & 0.896 & 0.696 & 0.650 & 0.125 & 0.004 & 0.000 & 0.000 \\
     & $\BLEU$ & 0.969 & 0.923 & 0.878 & 0.446 & 0.130 & 0.069 & 0.049 \\
     & $\chrFPP$ & 0.981 & 0.961 & 0.937 & 0.602 & 0.288 & 0.210 & 0.170 \\
    \midrule
    \multirow{4}{=}{SVO → SOV} & Exact Match & 0.121 & 0.121 & 0.121 & 0.013 & 0.000 & 0.000 & 0.000 \\
     & Bag of Words & 0.804 & 0.654 & 0.571 & 0.133 & 0.000 & 0.004 & 0.004 \\
     & $\BLEU$ & 0.504 & 0.428 & 0.414 & 0.172 & 0.055 & 0.063 & 0.047 \\
     & $\chrFPP$ & 0.840 & 0.786 & 0.792 & 0.514 & 0.213 & 0.231 & 0.168 \\
    \midrule
    \multirow{4}{=}{SVO → OVS} & Exact Match & 0.083 & 0.062 & 0.062 & 0.000 & 0.000 & 0.000 & 0.000 \\
     & Bag of Words & 0.796 & 0.821 & 0.729 & 0.083 & 0.004 & 0.004 & 0.000 \\
     & $\BLEU$ & 0.371 & 0.388 & 0.304 & 0.131 & 0.062 & 0.034 & 0.041 \\
     & $\chrFPP$ & 0.761 & 0.782 & 0.736 & 0.462 & 0.237 & 0.161 & 0.203 \\
    \bottomrule
  \end{tabularx}
  \caption{Mean results for \texttt{gpt-5-nano} in the word order experiment, grouped by target word-order condition and grammar size.}
  \label{tab:results-wordorder-grammar:gpt-5-nano}
\end{table*}

\begin{table*}[ht]
  \centering
  \small
  \sisetup{table-format=1.3}
  \begin{tabularx}{\textwidth}{>{\raggedright\arraybackslash}X l S S S S S S S}
    \toprule
    \textbf{Condition} & \textbf{Metric} & \textbf{46} & \textbf{66} & \textbf{106} & \textbf{826} & \textbf{4,026} & \textbf{6,026} & \textbf{8,026} \\
    \midrule
    \multirow{4}{=}{SVO → SVO} & Exact Match & 0.371 & 0.354 & 0.158 & 0.000 & 0.000 & 0.000 & 0.000 \\
     & Bag of Words & 0.392 & 0.362 & 0.171 & 0.000 & 0.000 & 0.000 & 0.000 \\
     & $\BLEU$ & 0.761 & 0.756 & 0.566 & 0.066 & 0.033 & 0.024 & 0.026 \\
     & $\chrFPP$ & 0.856 & 0.853 & 0.735 & 0.283 & 0.145 & 0.155 & 0.137 \\
    \midrule
    \multirow{4}{=}{SVO → SOV} & Exact Match & 0.000 & 0.000 & 0.000 & 0.000 & 0.000 & 0.000 & 0.000 \\
     & Bag of Words & 0.412 & 0.400 & 0.163 & 0.000 & 0.000 & 0.000 & 0.000 \\
     & $\BLEU$ & 0.368 & 0.303 & 0.217 & 0.053 & 0.028 & 0.027 & 0.027 \\
     & $\chrFPP$ & 0.740 & 0.706 & 0.592 & 0.245 & 0.126 & 0.141 & 0.118 \\
    \midrule
    \multirow{4}{=}{SVO → OVS} & Exact Match & 0.000 & 0.000 & 0.000 & 0.000 & 0.000 & 0.000 & 0.000 \\
     & Bag of Words & 0.263 & 0.342 & 0.150 & 0.000 & 0.000 & 0.000 & 0.000 \\
     & $\BLEU$ & 0.258 & 0.288 & 0.176 & 0.047 & 0.033 & 0.020 & 0.028 \\
     & $\chrFPP$ & 0.655 & 0.675 & 0.570 & 0.269 & 0.135 & 0.131 & 0.164 \\
    \bottomrule
  \end{tabularx}
  \caption{Mean results for \texttt{gemma-3-12b-it} in the word order experiment, grouped by target word-order condition and grammar size.}
  \label{tab:results-wordorder-grammar:google_gemma-3-12b-it}
\end{table*}

\begin{table*}[ht]
  \centering
  \small
  \sisetup{table-format=1.3}
  \begin{tabularx}{\textwidth}{>{\raggedright\arraybackslash}X l S S S S S S S}
    \toprule
    \textbf{Condition} & \textbf{Metric} & \textbf{46} & \textbf{66} & \textbf{106} & \textbf{826} & \textbf{4,026} & \textbf{6,026} & \textbf{8,026} \\
    \midrule
    \multirow{4}{=}{SVO → SVO} & Exact Match & 0.000 & 0.000 & 0.000 & 0.000 & 0.000 & 0.000 & 0.000 \\
     & Bag of Words & 0.000 & 0.000 & 0.000 & 0.000 & 0.000 & 0.000 & 0.000 \\
     & $\BLEU$ & 0.006 & 0.013 & 0.002 & 0.001 & 0.002 & 0.001 & 0.001 \\
     & $\chrFPP$ & 0.116 & 0.106 & 0.103 & 0.095 & 0.100 & 0.094 & 0.083 \\
    \midrule
    \multirow{4}{=}{SVO → SOV} & Exact Match & 0.000 & 0.000 & 0.000 & 0.000 & 0.000 & 0.000 & 0.000 \\
     & Bag of Words & 0.000 & 0.000 & 0.004 & 0.000 & 0.000 & 0.000 & 0.000 \\
     & $\BLEU$ & 0.006 & 0.008 & 0.006 & 0.001 & 0.003 & 0.001 & 0.002 \\
     & $\chrFPP$ & 0.115 & 0.115 & 0.114 & 0.105 & 0.097 & 0.093 & 0.090 \\
    \midrule
    \multirow{4}{=}{SVO → OVS} & Exact Match & 0.000 & 0.000 & 0.000 & 0.000 & 0.000 & 0.000 & 0.000 \\
     & Bag of Words & 0.000 & 0.000 & 0.000 & 0.000 & 0.000 & 0.000 & 0.000 \\
     & $\BLEU$ & 0.003 & 0.005 & 0.006 & 0.001 & 0.001 & 0.001 & 0.001 \\
     & $\chrFPP$ & 0.109 & 0.110 & 0.113 & 0.093 & 0.092 & 0.091 & 0.092 \\
    \bottomrule
  \end{tabularx}
  \caption{Mean results for \texttt{gemma-3-4b-it} in the word order experiment, grouped by target word-order condition and grammar size.}
  \label{tab:results-wordorder-grammar:google_gemma-3-4b-it}
\end{table*}

\begin{table*}[ht]
  \centering
  \small
  \sisetup{table-format=1.3}
  \begin{tabularx}{\textwidth}{>{\raggedright\arraybackslash}X l S S S S S S S}
    \toprule
    \textbf{Condition} & \textbf{Metric} & \textbf{46} & \textbf{66} & \textbf{106} & \textbf{826} & \textbf{4,026} & \textbf{6,026} & \textbf{8,026} \\
    \midrule
    \multirow{4}{=}{SVO → SVO} & Exact Match & 0.000 & 0.000 & 0.000 & 0.000 & \multicolumn{1}{c}{---} & \multicolumn{1}{c}{---} & \multicolumn{1}{c}{---} \\
     & Bag of Words & 0.000 & 0.000 & 0.000 & 0.000 & \multicolumn{1}{c}{---} & \multicolumn{1}{c}{---} & \multicolumn{1}{c}{---} \\
     & $\BLEU$ & 0.000 & 0.001 & 0.000 & 0.001 & \multicolumn{1}{c}{---} & \multicolumn{1}{c}{---} & \multicolumn{1}{c}{---} \\
     & $\chrFPP$ & 0.101 & 0.091 & 0.102 & 0.101 & \multicolumn{1}{c}{---} & \multicolumn{1}{c}{---} & \multicolumn{1}{c}{---} \\
    \midrule
    \multirow{4}{=}{SVO → SOV} & Exact Match & 0.000 & 0.000 & 0.000 & 0.000 & \multicolumn{1}{c}{---} & \multicolumn{1}{c}{---} & \multicolumn{1}{c}{---} \\
     & Bag of Words & 0.000 & 0.000 & 0.000 & 0.000 & \multicolumn{1}{c}{---} & \multicolumn{1}{c}{---} & \multicolumn{1}{c}{---} \\
     & $\BLEU$ & 0.000 & 0.003 & 0.001 & 0.001 & \multicolumn{1}{c}{---} & \multicolumn{1}{c}{---} & \multicolumn{1}{c}{---} \\
     & $\chrFPP$ & 0.102 & 0.105 & 0.106 & 0.110 & \multicolumn{1}{c}{---} & \multicolumn{1}{c}{---} & \multicolumn{1}{c}{---} \\
    \midrule
    \multirow{4}{=}{SVO → OVS} & Exact Match & 0.000 & 0.000 & 0.000 & 0.000 & \multicolumn{1}{c}{---} & \multicolumn{1}{c}{---} & \multicolumn{1}{c}{---} \\
     & Bag of Words & 0.000 & 0.000 & 0.000 & 0.000 & \multicolumn{1}{c}{---} & \multicolumn{1}{c}{---} & \multicolumn{1}{c}{---} \\
     & $\BLEU$ & 0.000 & 0.000 & 0.001 & 0.001 & \multicolumn{1}{c}{---} & \multicolumn{1}{c}{---} & \multicolumn{1}{c}{---} \\
     & $\chrFPP$ & 0.102 & 0.101 & 0.106 & 0.102 & \multicolumn{1}{c}{---} & \multicolumn{1}{c}{---} & \multicolumn{1}{c}{---} \\
    \bottomrule
  \end{tabularx}
  \caption{Mean results for \texttt{gemma-3-1b-it} in the word order experiment, grouped by target word-order condition and grammar size.}
  \label{tab:results-wordorder-grammar:google_gemma-3-1b-it}
\end{table*}
\begin{table*}[ht]
  \centering
  \small
  \sisetup{table-format=1.3}
  \begin{tabularx}{\textwidth}{>{\raggedright\arraybackslash}X l S S S S S}
    \toprule
    \textbf{Condition} & \textbf{Metric} & \textbf{6} & \textbf{10.5} & \textbf{15.5} & \textbf{20.5} & \textbf{32.5} \\
    \midrule
    \multirow{4}{=}{SVO → SVO} & Exact Match & 0.822 & 0.752 & 0.663 & 0.722 & 0.538 \\
     & Bag of Words & 0.822 & 0.752 & 0.663 & 0.722 & 0.538 \\
     & $\BLEU$ & 0.922 & 0.916 & 0.894 & 0.927 & 0.871 \\
     & $\chrFPP$ & 0.952 & 0.954 & 0.939 & 0.960 & 0.926 \\
    \midrule
    \multirow{4}{=}{SVO → SOV} & Exact Match & 0.827 & 0.733 & 0.670 & 0.633 & 0.565 \\
     & Bag of Words & 0.832 & 0.733 & 0.679 & 0.655 & 0.579 \\
     & $\BLEU$ & 0.909 & 0.895 & 0.886 & 0.873 & 0.839 \\
     & $\chrFPP$ & 0.943 & 0.939 & 0.936 & 0.930 & 0.910 \\
    \midrule
    \multirow{4}{=}{SVO → OVS} & Exact Match & 0.847 & 0.762 & 0.649 & 0.631 & 0.541 \\
     & Bag of Words & 0.853 & 0.762 & 0.664 & 0.649 & 0.547 \\
     & $\BLEU$ & 0.923 & 0.889 & 0.852 & 0.839 & 0.775 \\
     & $\chrFPP$ & 0.956 & 0.945 & 0.929 & 0.918 & 0.885 \\
    \bottomrule
  \end{tabularx}
  \caption{Mean results for \texttt{gpt-5} in the word order experiment, grouped by target word-order condition and input string length.}
  \label{tab:results-wordorder-length:gpt-5}
\end{table*}

\begin{table*}[ht]
  \centering
  \small
  \sisetup{table-format=1.3}
  \begin{tabularx}{\textwidth}{>{\raggedright\arraybackslash}X l S S S S S}
    \toprule
    \textbf{Condition} & \textbf{Metric} & \textbf{6} & \textbf{10.5} & \textbf{15.5} & \textbf{20.5} & \textbf{32.5} \\
    \midrule
    \multirow{4}{=}{SVO → SVO} & Exact Match & 0.611 & 0.578 & 0.500 & 0.557 & 0.447 \\
     & Bag of Words & 0.611 & 0.578 & 0.500 & 0.557 & 0.447 \\
     & $\BLEU$ & 0.755 & 0.774 & 0.708 & 0.771 & 0.664 \\
     & $\chrFPP$ & 0.832 & 0.859 & 0.820 & 0.861 & 0.784 \\
    \midrule
    \multirow{4}{=}{SVO → SOV} & Exact Match & 0.578 & 0.539 & 0.497 & 0.482 & 0.450 \\
     & Bag of Words & 0.601 & 0.570 & 0.527 & 0.507 & 0.483 \\
     & $\BLEU$ & 0.698 & 0.672 & 0.624 & 0.615 & 0.554 \\
     & $\chrFPP$ & 0.797 & 0.801 & 0.779 & 0.778 & 0.742 \\
    \midrule
    \multirow{4}{=}{SVO → OVS} & Exact Match & 0.625 & 0.539 & 0.481 & 0.512 & 0.442 \\
     & Bag of Words & 0.631 & 0.563 & 0.516 & 0.524 & 0.462 \\
     & $\BLEU$ & 0.728 & 0.644 & 0.575 & 0.606 & 0.530 \\
     & $\chrFPP$ & 0.809 & 0.781 & 0.745 & 0.764 & 0.705 \\
    \bottomrule
  \end{tabularx}
  \caption{Mean results for \texttt{gpt-5-mini} in the word order experiment, grouped by target word-order condition and input string length.}
  \label{tab:results-wordorder-length:gpt-5-mini}
\end{table*}

\begin{table*}[ht]
  \centering
  \small
  \sisetup{table-format=1.3}
  \begin{tabularx}{\textwidth}{>{\raggedright\arraybackslash}X l S S S S S}
    \toprule
    \textbf{Condition} & \textbf{Metric} & \textbf{6} & \textbf{10.5} & \textbf{15.5} & \textbf{20.5} & \textbf{32.5} \\
    \midrule
    \multirow{4}{=}{SVO → SVO} & Exact Match & 0.413 & 0.382 & 0.313 & 0.332 & 0.252 \\
     & Bag of Words & 0.413 & 0.385 & 0.313 & 0.332 & 0.252 \\
     & $\BLEU$ & 0.583 & 0.534 & 0.467 & 0.496 & 0.395 \\
     & $\chrFPP$ & 0.688 & 0.643 & 0.567 & 0.581 & 0.485 \\
    \midrule
    \multirow{4}{=}{SVO → SOV} & Exact Match & 0.223 & 0.027 & 0.003 & 0.000 & 0.000 \\
     & Bag of Words & 0.425 & 0.367 & 0.275 & 0.255 & 0.210 \\
     & $\BLEU$ & 0.398 & 0.258 & 0.188 & 0.177 & 0.165 \\
     & $\chrFPP$ & 0.591 & 0.544 & 0.491 & 0.465 & 0.422 \\
    \midrule
    \multirow{4}{=}{SVO → OVS} & Exact Match & 0.136 & 0.012 & 0.000 & 0.000 & 0.000 \\
     & Bag of Words & 0.475 & 0.364 & 0.330 & 0.348 & 0.225 \\
     & $\BLEU$ & 0.376 & 0.196 & 0.138 & 0.133 & 0.108 \\
     & $\chrFPP$ & 0.608 & 0.491 & 0.450 & 0.464 & 0.375 \\
    \bottomrule
  \end{tabularx}
  \caption{Mean results for \texttt{gpt-5-nano} in the word order experiment, grouped by target word-order condition and input string length.}
  \label{tab:results-wordorder-length:gpt-5-nano}
\end{table*}

\begin{table*}[ht]
  \centering
  \small
  \sisetup{table-format=1.3}
  \begin{tabularx}{\textwidth}{>{\raggedright\arraybackslash}X l S S S S S}
    \toprule
    \textbf{Condition} & \textbf{Metric} & \textbf{6} & \textbf{10.5} & \textbf{15.5} & \textbf{20.5} & \textbf{32.5} \\
    \midrule
    \multirow{4}{=}{SVO → SVO} & Exact Match & 0.184 & 0.137 & 0.115 & 0.129 & 0.066 \\
     & Bag of Words & 0.193 & 0.146 & 0.120 & 0.132 & 0.069 \\
     & $\BLEU$ & 0.335 & 0.332 & 0.310 & 0.339 & 0.278 \\
     & $\chrFPP$ & 0.464 & 0.470 & 0.437 & 0.473 & 0.417 \\
    \midrule
    \multirow{4}{=}{SVO → SOV} & Exact Match & 0.000 & 0.000 & 0.000 & 0.000 & 0.000 \\
     & Bag of Words & 0.201 & 0.164 & 0.129 & 0.109 & 0.081 \\
     & $\BLEU$ & 0.182 & 0.158 & 0.130 & 0.132 & 0.124 \\
     & $\chrFPP$ & 0.378 & 0.394 & 0.367 & 0.394 & 0.372 \\
    \midrule
    \multirow{4}{=}{SVO → OVS} & Exact Match & 0.000 & 0.000 & 0.000 & 0.000 & 0.000 \\
     & Bag of Words & 0.192 & 0.120 & 0.086 & 0.088 & 0.053 \\
     & $\BLEU$ & 0.189 & 0.133 & 0.098 & 0.103 & 0.084 \\
     & $\chrFPP$ & 0.404 & 0.385 & 0.355 & 0.373 & 0.341 \\
    \bottomrule
  \end{tabularx}
  \caption{Mean results for \texttt{gemma-3-12b-it} in the word order experiment, grouped by target word-order condition and input string length.}
  \label{tab:results-wordorder-length:google_gemma-3-12b-it}
\end{table*}

\begin{table*}[ht]
  \centering
  \small
  \sisetup{table-format=1.3}
  \begin{tabularx}{\textwidth}{>{\raggedright\arraybackslash}X l S S S S S}
    \toprule
    \textbf{Condition} & \textbf{Metric} & \textbf{6} & \textbf{10.5} & \textbf{15.5} & \textbf{20.5} & \textbf{32.5} \\
    \midrule
    \multirow{4}{=}{SVO → SVO} & Exact Match & 0.000 & 0.000 & 0.000 & 0.000 & 0.000 \\
     & Bag of Words & 0.000 & 0.000 & 0.000 & 0.000 & 0.000 \\
     & $\BLEU$ & 0.004 & 0.003 & 0.006 & 0.002 & 0.002 \\
     & $\chrFPP$ & 0.080 & 0.097 & 0.106 & 0.105 & 0.108 \\
    \midrule
    \multirow{4}{=}{SVO → SOV} & Exact Match & 0.000 & 0.000 & 0.000 & 0.000 & 0.000 \\
     & Bag of Words & 0.003 & 0.000 & 0.000 & 0.000 & 0.000 \\
     & $\BLEU$ & 0.008 & 0.004 & 0.002 & 0.002 & 0.003 \\
     & $\chrFPP$ & 0.089 & 0.105 & 0.106 & 0.110 & 0.112 \\
    \midrule
    \multirow{4}{=}{SVO → OVS} & Exact Match & 0.000 & 0.000 & 0.000 & 0.000 & 0.000 \\
     & Bag of Words & 0.000 & 0.000 & 0.000 & 0.000 & 0.000 \\
     & $\BLEU$ & 0.004 & 0.002 & 0.002 & 0.002 & 0.002 \\
     & $\chrFPP$ & 0.089 & 0.100 & 0.103 & 0.105 & 0.105 \\
    \bottomrule
  \end{tabularx}
  \caption{Mean results for \texttt{gemma-3-4b-it} in the word order experiment, grouped by target word-order condition and input string length.}
  \label{tab:results-wordorder-length:google_gemma-3-4b-it}
\end{table*}

\begin{table*}[ht]
  \centering
  \small
  \sisetup{table-format=1.3}
  \begin{tabularx}{\textwidth}{>{\raggedright\arraybackslash}X l S S S S S}
    \toprule
    \textbf{Condition} & \textbf{Metric} & \textbf{6} & \textbf{10.5} & \textbf{15.5} & \textbf{20.5} & \textbf{32.5} \\
    \midrule
    \multirow{4}{=}{SVO → SVO} & Exact Match & 0.000 & 0.000 & 0.000 & 0.000 & 0.000 \\
     & Bag of Words & 0.000 & 0.000 & 0.000 & 0.000 & 0.000 \\
     & $\BLEU$ & 0.000 & 0.001 & 0.001 & 0.001 & 0.001 \\
     & $\chrFPP$ & 0.077 & 0.095 & 0.104 & 0.107 & 0.111 \\
    \midrule
    \multirow{4}{=}{SVO → SOV} & Exact Match & 0.000 & 0.000 & 0.000 & 0.000 & 0.000 \\
     & Bag of Words & 0.000 & 0.000 & 0.000 & 0.000 & 0.000 \\
     & $\BLEU$ & 0.001 & 0.001 & 0.001 & 0.001 & 0.002 \\
     & $\chrFPP$ & 0.084 & 0.101 & 0.112 & 0.115 & 0.121 \\
    \midrule
    \multirow{4}{=}{SVO → OVS} & Exact Match & 0.000 & 0.000 & 0.000 & 0.000 & 0.000 \\
     & Bag of Words & 0.000 & 0.000 & 0.000 & 0.000 & 0.000 \\
     & $\BLEU$ & 0.000 & 0.000 & 0.001 & 0.001 & 0.001 \\
     & $\chrFPP$ & 0.080 & 0.101 & 0.107 & 0.112 & 0.115 \\
    \bottomrule
  \end{tabularx}
  \caption{Mean results for \texttt{gemma-3-1b-it} in the word order experiment, grouped by target word-order condition and input string length.}
  \label{tab:results-wordorder-length:google_gemma-3-1b-it}
\end{table*}

\subsection{Morphology Experiment}

\begin{figure}[ht]
    \centering
    \includegraphics[width=\linewidth]{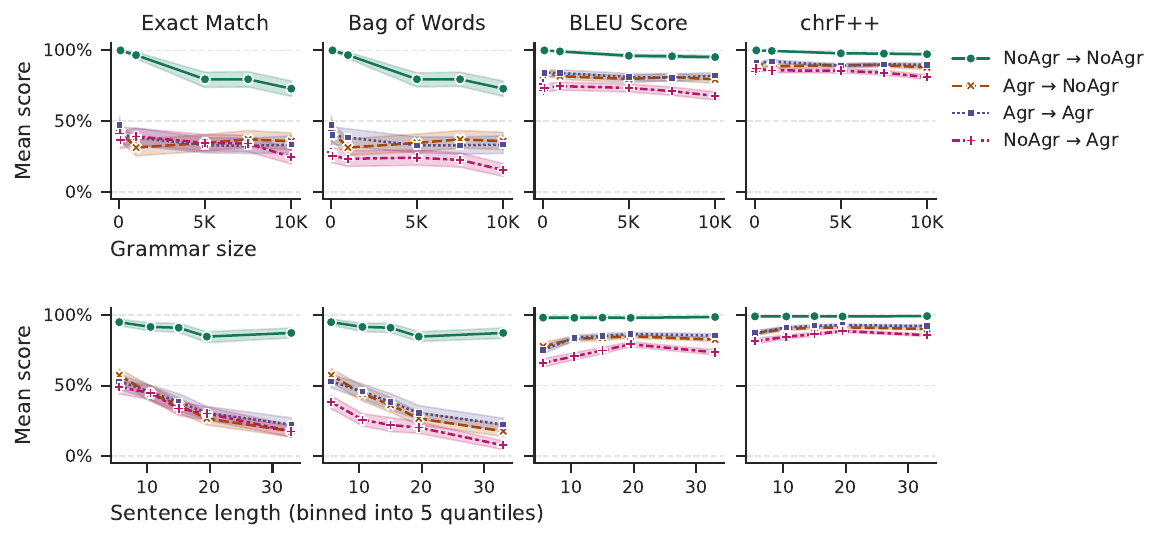}
    \caption{Full results for \texttt{gpt-5} on the morphology experiment. Error bars show 95\% confidence interval.}
    \label{fig:morphology-gpt-5}
\end{figure}

\begin{figure}[ht]
    \centering
    \includegraphics[width=\linewidth]{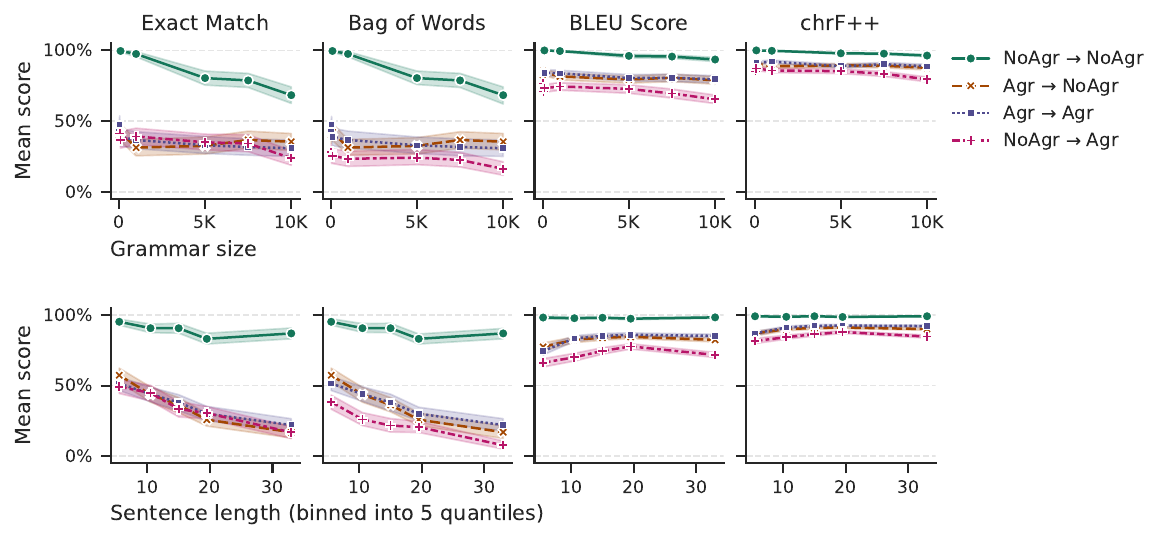}
    \caption{Full results for \texttt{gpt-5-mini} on the morphology experiment. Error bars show 95\% confidence interval.}
    \label{fig:morphology-gpt-5-mini}
\end{figure}

\begin{figure}[ht]
    \centering
    \includegraphics[width=\linewidth]{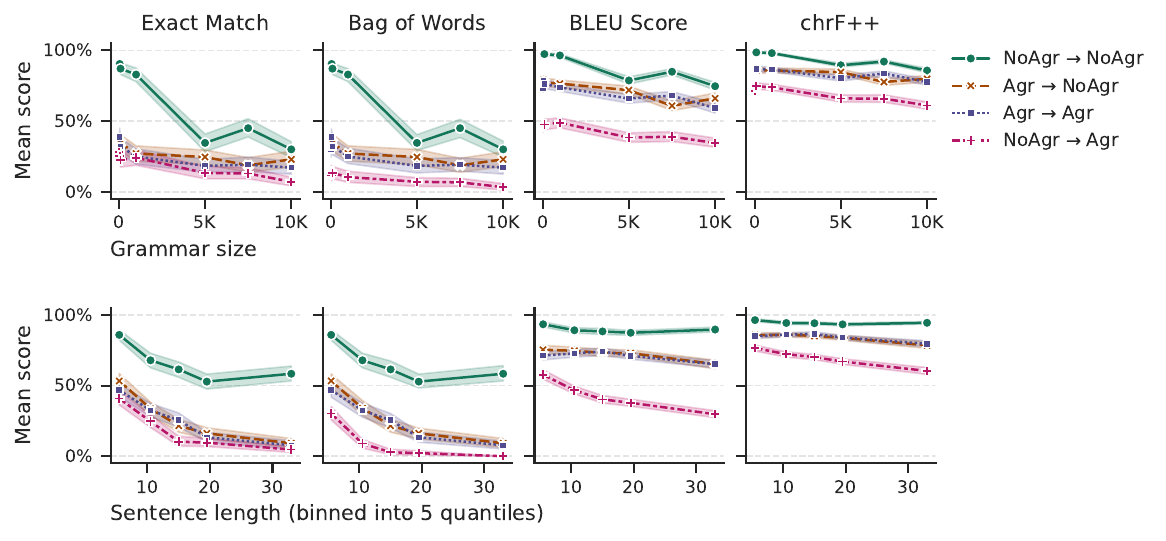}
    \caption{Full results for \texttt{gpt-5-nano} on the morphology experiment. Error bars show 95\% confidence interval.}
    \label{fig:morphology-gpt-5-nano}
\end{figure}

The figures and tables below show results on the word order experiment for \texttt{gpt-5} (\cref{fig:morphology-gpt-5,tab:results-agreement-grammar:gpt-5,tab:results-agreement-length:gpt-5}), \texttt{gpt-5-mini} (\cref{fig:morphology-gpt-5-mini,tab:results-agreement-grammar:gpt-5-mini,tab:results-agreement-length:gpt-5-mini}), and \texttt{gpt-5-nano} (\cref{fig:morphology-gpt-5-nano,tab:results-agreement-grammar:gpt-5-nano,tab:results-agreement-length:gpt-5-nano}).

\begin{table*}[ht]
  \centering
  \small
  \sisetup{table-format=1.3}
  \begin{tabularx}{\textwidth}{>{\raggedright\arraybackslash}X l S S S S S S S}
    \toprule
    \textbf{Condition} & \textbf{Metric} & \textbf{25} & \textbf{50} & \textbf{100} & \textbf{1,000} & \textbf{5,000} & \textbf{7,500} & \textbf{10,000} \\
    \midrule
    \multirow{4}{=}{NoAgr → NoAgr} & Exact Match & 0.996 & 1.000 & 1.000 & 0.967 & 0.796 & 0.796 & 0.729 \\
     & Bag of Words & 0.996 & 1.000 & 1.000 & 0.967 & 0.796 & 0.796 & 0.729 \\
     & $\BLEU$ & 0.999 & 1.000 & 1.000 & 0.993 & 0.962 & 0.958 & 0.953 \\
     & $\chrFPP$ & 1.000 & 1.000 & 1.000 & 0.996 & 0.979 & 0.977 & 0.973 \\
    \midrule
    \multirow{4}{=}{Agr → NoAgr} & Exact Match & 0.400 & 0.421 & 0.421 & 0.312 & 0.346 & 0.371 & 0.358 \\
     & Bag of Words & 0.400 & 0.421 & 0.421 & 0.312 & 0.346 & 0.371 & 0.358 \\
     & $\BLEU$ & 0.828 & 0.856 & 0.846 & 0.818 & 0.798 & 0.812 & 0.795 \\
     & $\chrFPP$ & 0.900 & 0.914 & 0.907 & 0.888 & 0.896 & 0.895 & 0.880 \\
    \midrule
    \multirow{4}{=}{Agr → Agr} & Exact Match & 0.446 & 0.471 & 0.400 & 0.383 & 0.329 & 0.325 & 0.333 \\
     & Bag of Words & 0.446 & 0.471 & 0.400 & 0.383 & 0.329 & 0.329 & 0.333 \\
     & $\BLEU$ & 0.842 & 0.840 & 0.841 & 0.840 & 0.812 & 0.806 & 0.821 \\
     & $\chrFPP$ & 0.925 & 0.914 & 0.911 & 0.922 & 0.891 & 0.902 & 0.898 \\
    \midrule
    \multirow{4}{=}{NoAgr → Agr} & Exact Match & 0.408 & 0.412 & 0.367 & 0.392 & 0.346 & 0.342 & 0.246 \\
     & Bag of Words & 0.283 & 0.258 & 0.254 & 0.233 & 0.242 & 0.225 & 0.154 \\
     & $\BLEU$ & 0.765 & 0.718 & 0.734 & 0.747 & 0.734 & 0.712 & 0.678 \\
     & $\chrFPP$ & 0.867 & 0.857 & 0.872 & 0.858 & 0.856 & 0.842 & 0.811 \\
    \bottomrule
  \end{tabularx}
  \caption{Mean results for \texttt{gpt-5} in the morphology experiment, grouped by agreement condition and grammar size.}
  \label{tab:results-agreement-grammar:gpt-5}
\end{table*}

\begin{table*}[ht]
  \centering
  \small
  \sisetup{table-format=1.3}
  \begin{tabularx}{\textwidth}{>{\raggedright\arraybackslash}X l S S S S S S S}
    \toprule
    \textbf{Condition} & \textbf{Metric} & \textbf{25} & \textbf{50} & \textbf{100} & \textbf{1,000} & \textbf{5,000} & \textbf{7,500} & \textbf{10,000} \\
    \midrule
    \multirow{4}{=}{NoAgr → NoAgr} & Exact Match & 0.996 & 1.000 & 0.996 & 0.975 & 0.804 & 0.787 & 0.683 \\
     & Bag of Words & 0.996 & 1.000 & 0.996 & 0.975 & 0.804 & 0.787 & 0.683 \\
     & $\BLEU$ & 0.999 & 1.000 & 1.000 & 0.995 & 0.961 & 0.956 & 0.936 \\
     & $\chrFPP$ & 1.000 & 1.000 & 1.000 & 0.997 & 0.979 & 0.977 & 0.963 \\
    \midrule
    \multirow{4}{=}{Agr → NoAgr} & Exact Match & 0.400 & 0.421 & 0.421 & 0.312 & 0.325 & 0.367 & 0.354 \\
     & Bag of Words & 0.400 & 0.421 & 0.421 & 0.312 & 0.325 & 0.367 & 0.354 \\
     & $\BLEU$ & 0.828 & 0.856 & 0.844 & 0.816 & 0.795 & 0.806 & 0.789 \\
     & $\chrFPP$ & 0.900 & 0.915 & 0.906 & 0.887 & 0.894 & 0.891 & 0.877 \\
    \midrule
    \multirow{4}{=}{Agr → Agr} & Exact Match & 0.442 & 0.475 & 0.388 & 0.367 & 0.329 & 0.317 & 0.308 \\
     & Bag of Words & 0.442 & 0.475 & 0.388 & 0.367 & 0.329 & 0.317 & 0.308 \\
     & $\BLEU$ & 0.841 & 0.845 & 0.841 & 0.838 & 0.805 & 0.806 & 0.796 \\
     & $\chrFPP$ & 0.925 & 0.917 & 0.912 & 0.920 & 0.888 & 0.902 & 0.887 \\
    \midrule
    \multirow{4}{=}{NoAgr → Agr} & Exact Match & 0.408 & 0.412 & 0.367 & 0.392 & 0.350 & 0.342 & 0.237 \\
     & Bag of Words & 0.283 & 0.254 & 0.254 & 0.233 & 0.242 & 0.225 & 0.163 \\
     & $\BLEU$ & 0.766 & 0.709 & 0.733 & 0.745 & 0.727 & 0.695 & 0.654 \\
     & $\chrFPP$ & 0.868 & 0.857 & 0.872 & 0.858 & 0.853 & 0.835 & 0.795 \\
    \bottomrule
  \end{tabularx}
  \caption{Mean results for \texttt{gpt-5-mini} in the morphology experiment, grouped by agreement condition and grammar size.}
  \label{tab:results-agreement-grammar:gpt-5-mini}
\end{table*}

\begin{table*}[ht]
  \centering
  \small
  \sisetup{table-format=1.3}
  \begin{tabularx}{\textwidth}{>{\raggedright\arraybackslash}X l S S S S S S S}
    \toprule
    \textbf{Condition} & \textbf{Metric} & \textbf{25} & \textbf{50} & \textbf{100} & \textbf{1,000} & \textbf{5,000} & \textbf{7,500} & \textbf{10,000} \\
    \midrule
    \multirow{4}{=}{NoAgr → NoAgr} & Exact Match & 0.887 & 0.904 & 0.871 & 0.829 & 0.346 & 0.450 & 0.300 \\
     & Bag of Words & 0.887 & 0.904 & 0.871 & 0.829 & 0.346 & 0.450 & 0.300 \\
     & $\BLEU$ & 0.970 & 0.980 & 0.974 & 0.964 & 0.787 & 0.849 & 0.747 \\
     & $\chrFPP$ & 0.984 & 0.988 & 0.986 & 0.981 & 0.895 & 0.921 & 0.857 \\
    \midrule
    \multirow{4}{=}{Agr → NoAgr} & Exact Match & 0.346 & 0.346 & 0.346 & 0.271 & 0.246 & 0.188 & 0.229 \\
     & Bag of Words & 0.346 & 0.350 & 0.346 & 0.271 & 0.246 & 0.188 & 0.229 \\
     & $\BLEU$ & 0.759 & 0.790 & 0.773 & 0.765 & 0.718 & 0.608 & 0.661 \\
     & $\chrFPP$ & 0.860 & 0.879 & 0.858 & 0.858 & 0.846 & 0.777 & 0.799 \\
    \midrule
    \multirow{4}{=}{Agr → Agr} & Exact Match & 0.300 & 0.388 & 0.321 & 0.250 & 0.183 & 0.192 & 0.171 \\
     & Bag of Words & 0.300 & 0.388 & 0.321 & 0.250 & 0.183 & 0.192 & 0.171 \\
     & $\BLEU$ & 0.734 & 0.783 & 0.764 & 0.738 & 0.660 & 0.680 & 0.592 \\
     & $\chrFPP$ & 0.863 & 0.876 & 0.868 & 0.865 & 0.804 & 0.836 & 0.776 \\
    \midrule
    \multirow{4}{=}{NoAgr → Agr} & Exact Match & 0.279 & 0.254 & 0.225 & 0.242 & 0.133 & 0.129 & 0.071 \\
     & Bag of Words & 0.142 & 0.121 & 0.133 & 0.104 & 0.071 & 0.067 & 0.033 \\
     & $\BLEU$ & 0.464 & 0.470 & 0.477 & 0.488 & 0.383 & 0.390 & 0.345 \\
     & $\chrFPP$ & 0.715 & 0.739 & 0.747 & 0.739 & 0.660 & 0.658 & 0.611 \\
    \bottomrule
  \end{tabularx}
  \caption{Mean results for \texttt{gpt-5-nano} in the morphology experiment, grouped by agreement condition and grammar size.}
  \label{tab:results-agreement-grammar:gpt-5-nano}
\end{table*}
\begin{table*}[ht]
  \centering
  \small
  \sisetup{table-format=1.3}
  \begin{tabularx}{\textwidth}{>{\raggedright\arraybackslash}X l S S S S S}
    \toprule
    \textbf{Condition} & \textbf{Metric} & \textbf{5.5} & \textbf{10.5} & \textbf{15} & \textbf{19.5} & \textbf{33} \\
    \midrule
    \multirow{4}{=}{NoAgr → NoAgr} & Exact Match & 0.948 & 0.914 & 0.909 & 0.846 & 0.872 \\
     & Bag of Words & 0.948 & 0.914 & 0.909 & 0.846 & 0.872 \\
     & $\BLEU$ & 0.981 & 0.979 & 0.980 & 0.979 & 0.985 \\
     & $\chrFPP$ & 0.989 & 0.988 & 0.990 & 0.988 & 0.991 \\
    \midrule
    \multirow{4}{=}{Agr → NoAgr} & Exact Match & 0.573 & 0.442 & 0.363 & 0.267 & 0.177 \\
     & Bag of Words & 0.573 & 0.442 & 0.363 & 0.267 & 0.177 \\
     & $\BLEU$ & 0.777 & 0.830 & 0.838 & 0.848 & 0.825 \\
     & $\chrFPP$ & 0.870 & 0.900 & 0.908 & 0.912 & 0.901 \\
    \midrule
    \multirow{4}{=}{Agr → Agr} & Exact Match & 0.530 & 0.454 & 0.384 & 0.306 & 0.224 \\
     & Bag of Words & 0.530 & 0.457 & 0.384 & 0.306 & 0.224 \\
     & $\BLEU$ & 0.752 & 0.837 & 0.851 & 0.864 & 0.854 \\
     & $\chrFPP$ & 0.871 & 0.911 & 0.922 & 0.927 & 0.921 \\
    \midrule
    \multirow{4}{=}{NoAgr → Agr} & Exact Match & 0.489 & 0.447 & 0.338 & 0.302 & 0.176 \\
     & Bag of Words & 0.384 & 0.257 & 0.221 & 0.203 & 0.078 \\
     & $\BLEU$ & 0.660 & 0.707 & 0.749 & 0.792 & 0.735 \\
     & $\chrFPP$ & 0.814 & 0.844 & 0.863 & 0.886 & 0.856 \\
    \bottomrule
  \end{tabularx}
  \caption{Mean results for \texttt{gpt-5} in the morphology experiment, grouped by agreement condition and input string length.}
  \label{tab:results-agreement-length:gpt-5}
\end{table*}

\begin{table*}[ht]
  \centering
  \small
  \sisetup{table-format=1.3}
  \begin{tabularx}{\textwidth}{>{\raggedright\arraybackslash}X l S S S S S}
    \toprule
    \textbf{Condition} & \textbf{Metric} & \textbf{5.5} & \textbf{10.5} & \textbf{15} & \textbf{19.5} & \textbf{33} \\
    \midrule
    \multirow{4}{=}{NoAgr → NoAgr} & Exact Match & 0.951 & 0.906 & 0.905 & 0.830 & 0.869 \\
     & Bag of Words & 0.951 & 0.906 & 0.905 & 0.830 & 0.869 \\
     & $\BLEU$ & 0.981 & 0.976 & 0.980 & 0.973 & 0.982 \\
     & $\chrFPP$ & 0.990 & 0.986 & 0.990 & 0.985 & 0.990 \\
    \midrule
    \multirow{4}{=}{Agr → NoAgr} & Exact Match & 0.573 & 0.436 & 0.363 & 0.259 & 0.170 \\
     & Bag of Words & 0.573 & 0.436 & 0.363 & 0.259 & 0.170 \\
     & $\BLEU$ & 0.774 & 0.825 & 0.838 & 0.845 & 0.823 \\
     & $\chrFPP$ & 0.869 & 0.899 & 0.907 & 0.910 & 0.900 \\
    \midrule
    \multirow{4}{=}{Agr → Agr} & Exact Match & 0.514 & 0.440 & 0.380 & 0.300 & 0.221 \\
     & Bag of Words & 0.514 & 0.440 & 0.380 & 0.300 & 0.221 \\
     & $\BLEU$ & 0.744 & 0.833 & 0.851 & 0.859 & 0.851 \\
     & $\chrFPP$ & 0.868 & 0.910 & 0.922 & 0.924 & 0.921 \\
    \midrule
    \multirow{4}{=}{NoAgr → Agr} & Exact Match & 0.492 & 0.447 & 0.335 & 0.305 & 0.169 \\
     & Bag of Words & 0.384 & 0.260 & 0.217 & 0.206 & 0.078 \\
     & $\BLEU$ & 0.662 & 0.700 & 0.745 & 0.777 & 0.717 \\
     & $\chrFPP$ & 0.815 & 0.842 & 0.863 & 0.878 & 0.846 \\
    \bottomrule
  \end{tabularx}
  \caption{Mean results for \texttt{gpt-5-mini} in the morphology experiment, grouped by agreement condition and input string length.}
  \label{tab:results-agreement-length:gpt-5-mini}
\end{table*}

\begin{table*}[ht]
  \centering
  \small
  \sisetup{table-format=1.3}
  \begin{tabularx}{\textwidth}{>{\raggedright\arraybackslash}X l S S S S S}
    \toprule
    \textbf{Condition} & \textbf{Metric} & \textbf{5.5} & \textbf{10.5} & \textbf{15} & \textbf{19.5} & \textbf{33} \\
    \midrule
    \multirow{4}{=}{NoAgr → NoAgr} & Exact Match & 0.858 & 0.678 & 0.614 & 0.528 & 0.583 \\
     & Bag of Words & 0.858 & 0.678 & 0.614 & 0.528 & 0.583 \\
     & $\BLEU$ & 0.933 & 0.891 & 0.883 & 0.873 & 0.896 \\
     & $\chrFPP$ & 0.963 & 0.942 & 0.941 & 0.933 & 0.944 \\
    \midrule
    \multirow{4}{=}{Agr → NoAgr} & Exact Match & 0.533 & 0.336 & 0.216 & 0.162 & 0.092 \\
     & Bag of Words & 0.533 & 0.339 & 0.216 & 0.162 & 0.092 \\
     & $\BLEU$ & 0.753 & 0.746 & 0.731 & 0.729 & 0.652 \\
     & $\chrFPP$ & 0.856 & 0.859 & 0.851 & 0.838 & 0.784 \\
    \midrule
    \multirow{4}{=}{Agr → Agr} & Exact Match & 0.467 & 0.325 & 0.255 & 0.134 & 0.078 \\
     & Bag of Words & 0.467 & 0.325 & 0.255 & 0.134 & 0.078 \\
     & $\BLEU$ & 0.712 & 0.728 & 0.741 & 0.710 & 0.651 \\
     & $\chrFPP$ & 0.849 & 0.860 & 0.864 & 0.840 & 0.795 \\
    \midrule
    \multirow{4}{=}{NoAgr → Agr} & Exact Match & 0.408 & 0.247 & 0.103 & 0.096 & 0.047 \\
     & Bag of Words & 0.303 & 0.089 & 0.028 & 0.022 & 0.000 \\
     & $\BLEU$ & 0.577 & 0.466 & 0.402 & 0.378 & 0.297 \\
     & $\chrFPP$ & 0.765 & 0.721 & 0.701 & 0.669 & 0.603 \\
    \bottomrule
  \end{tabularx}
  \caption{Mean results for \texttt{gpt-5-nano} in the morphology experiment, grouped by agreement condition and input string length.}
  \label{tab:results-agreement-length:gpt-5-nano}
\end{table*}

\subsection{Orthography Experiment}

\begin{figure}[ht]
    \centering
    \includegraphics[width=\linewidth]{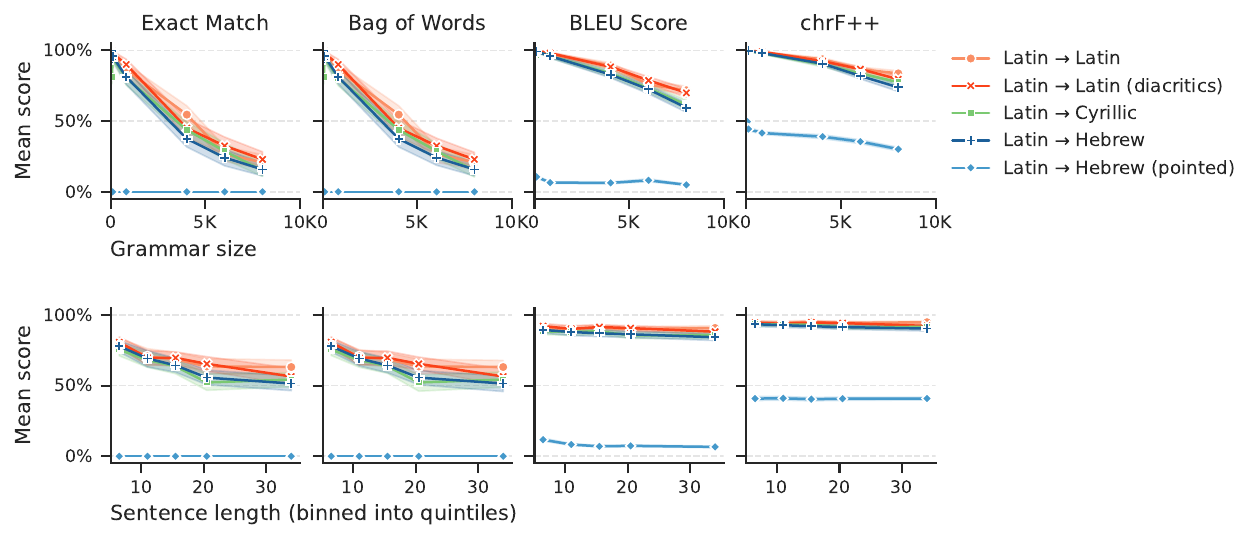}
    \caption{Full results for \texttt{gpt-5} on the orthography experiment. Error bars show 95\% confidence interval.}
    \label{fig:orthography-gpt-5}
\end{figure}

\begin{figure}[ht]
    \centering
    \includegraphics[width=\linewidth]{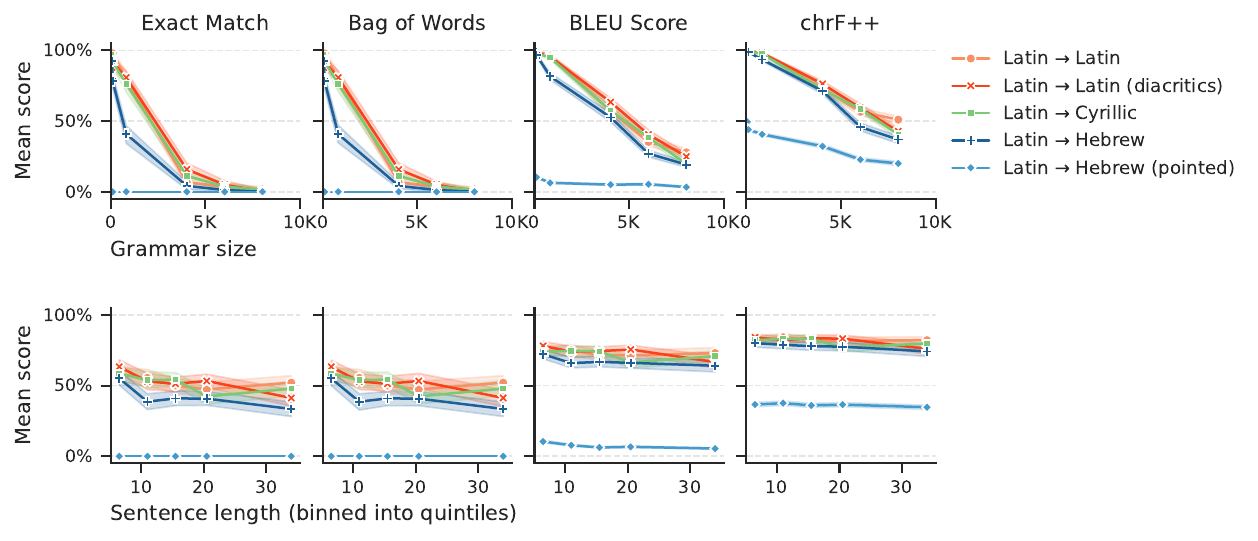}
    \caption{Full results for \texttt{gpt-5-mini} on the orthography experiment. Error bars show 95\% confidence interval.}
    \label{fig:orthography-gpt-5-mini}
\end{figure}

\begin{figure}[ht]
    \centering
    \includegraphics[width=\linewidth]{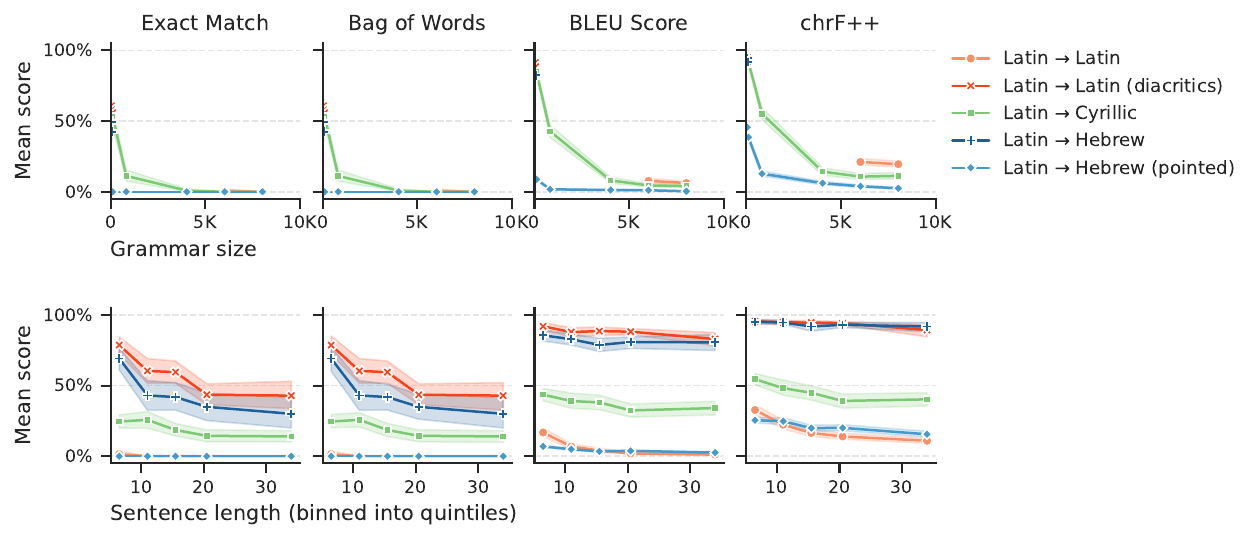}
    \caption{Full results for \texttt{gpt-5-nano} on the orthography experiment. Error bars show 95\% confidence interval.}
    \label{fig:orthography-gpt-5-nano}
\end{figure}

\begin{figure}[ht]
    \centering
    \includegraphics[width=\linewidth]{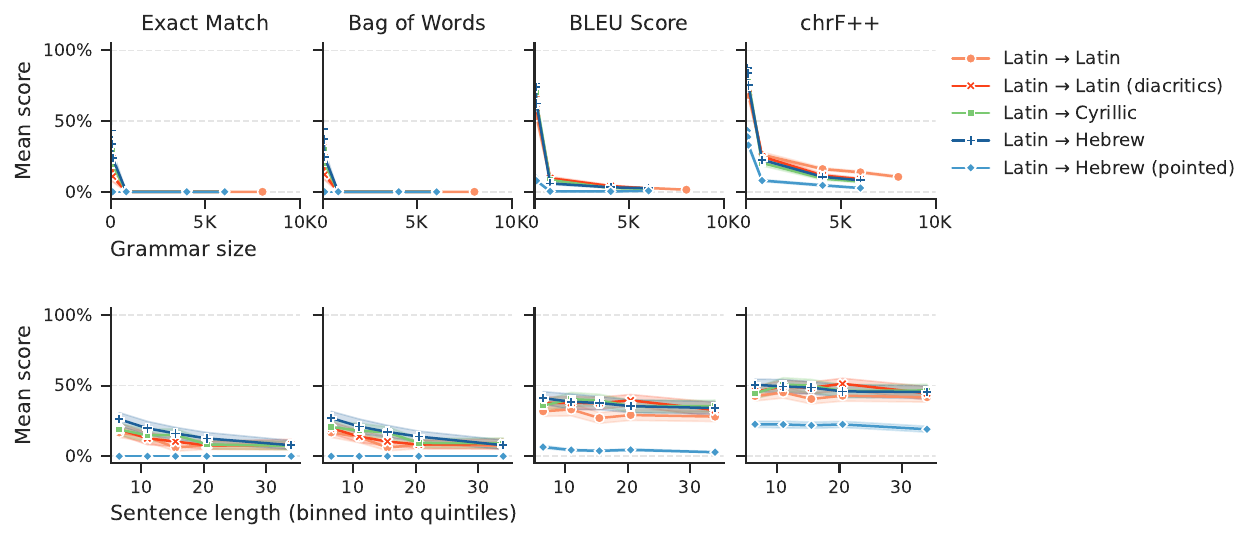}
    \caption{Full results for \texttt{gemma-3-12b-it} on the orthography experiment. Error bars show 95\% confidence interval.}
    \label{fig:orthography-gemma-3-12b-it}
\end{figure}

\begin{figure}[ht]
    \centering
    \includegraphics[width=\linewidth]{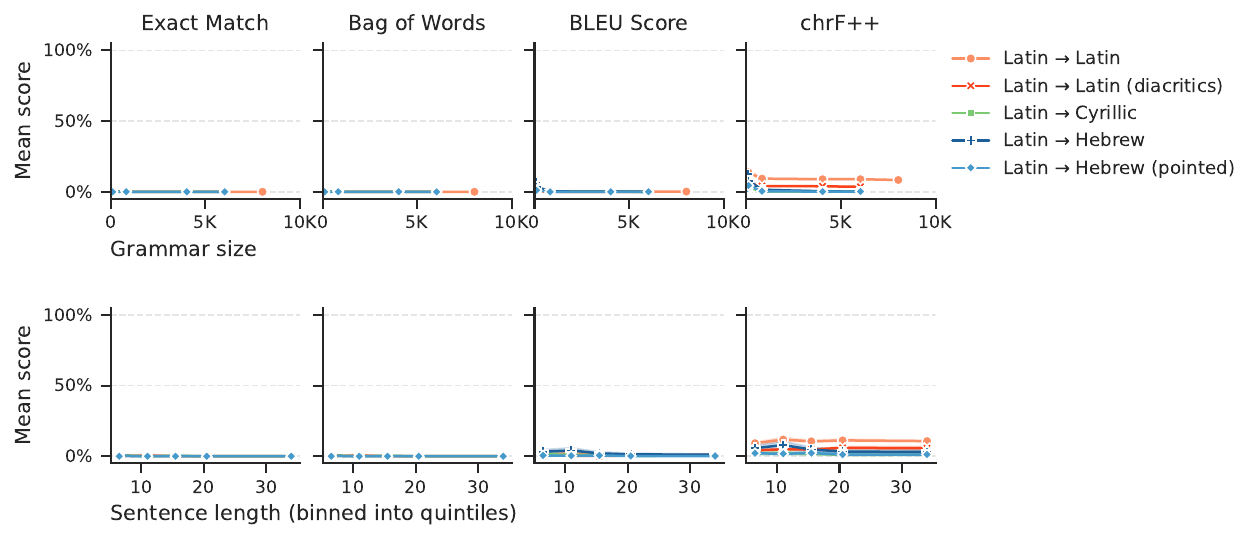}
    \caption{Full results for \texttt{gemma-3-4b-it} on the orthography experiment. Error bars show 95\% confidence interval.}
    \label{fig:orthography-gemma-3-4b-it}
\end{figure}

\begin{figure}[ht]
    \centering
    \includegraphics[width=\linewidth]{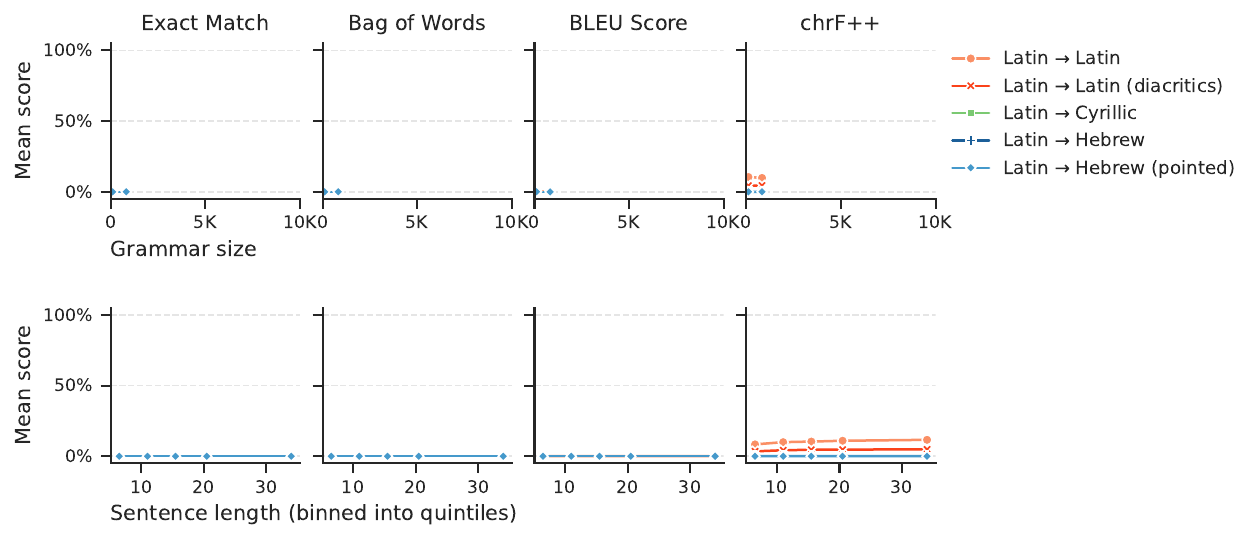}
    \caption{Full results for \texttt{gemma-3-1b-it} on the orthography experiment. Error bars show 95\% confidence interval.}
    \label{fig:orthography-gemma-3-1b-it}
\end{figure}

The figures and tables below show results on the word order experiment for \texttt{gpt-5} (\cref{fig:orthography-gpt-5,tab:results-orthography-grammar:gpt-5,tab:results-orthography-length:gpt-5}), \texttt{gpt-5-mini} (\cref{fig:orthography-gpt-5-mini,tab:results-orthography-grammar:gpt-5-mini,tab:results-orthography-length:gpt-5-mini}), \texttt{gpt-5-nano} (\cref{fig:orthography-gpt-5-nano,tab:results-orthography-grammar:gpt-5-nano,tab:results-orthography-length:gpt-5-nano}), \texttt{gemma-3-12b-it} (\cref{fig:orthography-gemma-3-12b-it,tab:results-orthography-grammar:google_gemma-3-12b-it,tab:results-orthography-length:google_gemma-3-12b-it}), \texttt{gemma-3-4b-it} (\cref{fig:orthography-gemma-3-4b-it,tab:results-orthography-grammar:google_gemma-3-4b-it,tab:results-orthography-length:google_gemma-3-4b-it}), and \texttt{gemma-3-1b-it} (\cref{fig:orthography-gemma-3-1b-it,tab:results-orthography-grammar:google_gemma-3-1b-it,tab:results-orthography-length:google_gemma-3-1b-it}). Note that \texttt{gemma-3-1b-it} has a much shorter maximum context window which precludes us from evaluating it on larger grammars.

\begin{table*}[ht]
  \centering
  \small
  \sisetup{table-format=1.3}
  \begin{tabularx}{\textwidth}{>{\raggedright\arraybackslash}X l S S S S S S S}
    \toprule
    \textbf{Condition} & \textbf{Metric} & \textbf{46} & \textbf{66} & \textbf{106} & \textbf{826} & \textbf{4,026} & \textbf{6,026} & \textbf{8,026} \\
    \midrule
    \multirow{4}{=}{Latin} & Exact Match & 0.992 & 0.983 & 1.000 & 0.863 & 0.546 & 0.267 & 0.208 \\
     & Bag of Words & 0.992 & 0.983 & 1.000 & 0.863 & 0.546 & 0.267 & 0.208 \\
     & $\BLEU$ & 0.999 & 0.994 & 1.000 & 0.973 & 0.893 & 0.758 & 0.718 \\
     & $\chrFPP$ & 1.000 & 0.997 & 1.000 & 0.983 & 0.937 & 0.866 & 0.837 \\
    \midrule
    \multirow{4}{=}{Latin + diac.} & Exact Match & 0.983 & 0.983 & 0.979 & 0.900 & 0.450 & 0.325 & 0.229 \\
     & Bag of Words & 0.983 & 0.983 & 0.979 & 0.900 & 0.450 & 0.325 & 0.229 \\
     & $\BLEU$ & 0.998 & 0.998 & 0.997 & 0.980 & 0.883 & 0.788 & 0.700 \\
     & $\chrFPP$ & 0.999 & 0.999 & 0.999 & 0.990 & 0.924 & 0.867 & 0.796 \\
    \midrule
    \multirow{4}{=}{Cyrillic} & Exact Match & 0.983 & 0.812 & 0.946 & 0.804 & 0.438 & 0.287 & 0.150 \\
     & Bag of Words & 0.983 & 0.812 & 0.946 & 0.804 & 0.438 & 0.287 & 0.150 \\
     & $\BLEU$ & 0.996 & 0.969 & 0.994 & 0.959 & 0.837 & 0.733 & 0.616 \\
     & $\chrFPP$ & 0.999 & 0.989 & 0.996 & 0.979 & 0.900 & 0.837 & 0.775 \\
    \midrule
    \multirow{4}{=}{Hebrew} & Exact Match & 0.996 & 0.938 & 0.958 & 0.812 & 0.375 & 0.242 & 0.158 \\
     & Bag of Words & 0.996 & 0.938 & 0.958 & 0.812 & 0.375 & 0.242 & 0.158 \\
     & $\BLEU$ & 1.000 & 0.990 & 0.993 & 0.959 & 0.828 & 0.724 & 0.595 \\
     & $\chrFPP$ & 1.000 & 0.996 & 0.998 & 0.981 & 0.906 & 0.819 & 0.738 \\
    \midrule
    \multirow{4}{=}{Hebrew + points} & Exact Match & 0.000 & 0.000 & 0.000 & 0.000 & 0.000 & 0.000 & 0.000 \\
     & Bag of Words & 0.000 & 0.000 & 0.000 & 0.000 & 0.000 & 0.000 & 0.000 \\
     & $\BLEU$ & 0.123 & 0.096 & 0.106 & 0.065 & 0.063 & 0.081 & 0.050 \\
     & $\chrFPP$ & 0.496 & 0.447 & 0.444 & 0.416 & 0.390 & 0.355 & 0.302 \\
    \bottomrule
  \end{tabularx}
  \caption{Mean results for \texttt{gpt-5} in the orthography experiment, grouped by target orthography and grammar size.}
  \label{tab:results-orthography-grammar:gpt-5}
\end{table*}

\begin{table*}[ht]
  \centering
  \small
  \sisetup{table-format=1.3}
  \begin{tabularx}{\textwidth}{>{\raggedright\arraybackslash}X l S S S S S S S}
    \toprule
    \textbf{Condition} & \textbf{Metric} & \textbf{46} & \textbf{66} & \textbf{106} & \textbf{826} & \textbf{4,026} & \textbf{6,026} & \textbf{8,026} \\
    \midrule
    \multirow{4}{=}{Latin} & Exact Match & 0.800 & 0.992 & 0.979 & 0.800 & 0.067 & 0.037 & 0.017 \\
     & Bag of Words & 0.800 & 0.992 & 0.979 & 0.800 & 0.067 & 0.037 & 0.017 \\
     & $\BLEU$ & 0.955 & 0.999 & 0.996 & 0.963 & 0.570 & 0.352 & 0.279 \\
     & $\chrFPP$ & 0.987 & 0.999 & 0.998 & 0.978 & 0.729 & 0.565 & 0.510 \\
    \midrule
    \multirow{4}{=}{Latin + diac.} & Exact Match & 0.917 & 0.879 & 0.917 & 0.808 & 0.158 & 0.050 & 0.004 \\
     & Bag of Words & 0.917 & 0.879 & 0.917 & 0.808 & 0.158 & 0.050 & 0.004 \\
     & $\BLEU$ & 0.984 & 0.984 & 0.986 & 0.956 & 0.634 & 0.405 & 0.248 \\
     & $\chrFPP$ & 0.995 & 0.995 & 0.995 & 0.981 & 0.763 & 0.593 & 0.428 \\
    \midrule
    \multirow{4}{=}{Cyrillic} & Exact Match & 0.967 & 0.804 & 0.921 & 0.762 & 0.113 & 0.037 & 0.013 \\
     & Bag of Words & 0.967 & 0.804 & 0.921 & 0.762 & 0.113 & 0.037 & 0.013 \\
     & $\BLEU$ & 0.994 & 0.960 & 0.982 & 0.949 & 0.573 & 0.383 & 0.199 \\
     & $\chrFPP$ & 0.998 & 0.986 & 0.994 & 0.975 & 0.724 & 0.585 & 0.407 \\
    \midrule
    \multirow{4}{=}{Hebrew} & Exact Match & 0.796 & 0.925 & 0.783 & 0.408 & 0.042 & 0.013 & 0.000 \\
     & Bag of Words & 0.796 & 0.925 & 0.783 & 0.408 & 0.042 & 0.013 & 0.000 \\
     & $\BLEU$ & 0.945 & 0.985 & 0.967 & 0.816 & 0.526 & 0.270 & 0.192 \\
     & $\chrFPP$ & 0.986 & 0.992 & 0.988 & 0.936 & 0.712 & 0.458 & 0.369 \\
    \midrule
    \multirow{4}{=}{Hebrew + points} & Exact Match & 0.000 & 0.000 & 0.000 & 0.000 & 0.000 & 0.000 & 0.000 \\
     & Bag of Words & 0.000 & 0.000 & 0.000 & 0.000 & 0.000 & 0.000 & 0.000 \\
     & $\BLEU$ & 0.120 & 0.094 & 0.103 & 0.064 & 0.050 & 0.053 & 0.033 \\
     & $\chrFPP$ & 0.494 & 0.444 & 0.440 & 0.406 & 0.322 & 0.228 & 0.200 \\
    \bottomrule
  \end{tabularx}
  \caption{Mean results for \texttt{gpt-5-mini} in the orthography experiment, grouped by target orthography and grammar size.}
  \label{tab:results-orthography-grammar:gpt-5-mini}
\end{table*}

\begin{table*}[ht]
  \centering
  \small
  \sisetup{table-format=1.3}
  \begin{tabularx}{\textwidth}{>{\raggedright\arraybackslash}X l S S S S S S S}
    \toprule
    \textbf{Condition} & \textbf{Metric} & \textbf{46} & \textbf{66} & \textbf{106} & \textbf{826} & \textbf{4,026} & \textbf{6,026} & \textbf{8,026} \\
    \midrule
    \multirow{4}{=}{Latin} & Exact Match & \multicolumn{1}{c}{---} & \multicolumn{1}{c}{---} & \multicolumn{1}{c}{---} & \multicolumn{1}{c}{---} & \multicolumn{1}{c}{---} & 0.008 & 0.000 \\
     & Bag of Words & \multicolumn{1}{c}{---} & \multicolumn{1}{c}{---} & \multicolumn{1}{c}{---} & \multicolumn{1}{c}{---} & \multicolumn{1}{c}{---} & 0.008 & 0.000 \\
     & $\BLEU$ & \multicolumn{1}{c}{---} & \multicolumn{1}{c}{---} & \multicolumn{1}{c}{---} & \multicolumn{1}{c}{---} & \multicolumn{1}{c}{---} & 0.077 & 0.062 \\
     & $\chrFPP$ & \multicolumn{1}{c}{---} & \multicolumn{1}{c}{---} & \multicolumn{1}{c}{---} & \multicolumn{1}{c}{---} & \multicolumn{1}{c}{---} & 0.211 & 0.195 \\
    \midrule
    \multirow{4}{=}{Latin + diac.} & Exact Match & 0.608 & 0.579 & 0.558 & \multicolumn{1}{c}{---} & \multicolumn{1}{c}{---} & \multicolumn{1}{c}{---} & \multicolumn{1}{c}{---} \\
     & Bag of Words & 0.608 & 0.579 & 0.558 & \multicolumn{1}{c}{---} & \multicolumn{1}{c}{---} & \multicolumn{1}{c}{---} & \multicolumn{1}{c}{---} \\
     & $\BLEU$ & 0.913 & 0.884 & 0.853 & \multicolumn{1}{c}{---} & \multicolumn{1}{c}{---} & \multicolumn{1}{c}{---} & \multicolumn{1}{c}{---} \\
     & $\chrFPP$ & 0.957 & 0.948 & 0.912 & \multicolumn{1}{c}{---} & \multicolumn{1}{c}{---} & \multicolumn{1}{c}{---} & \multicolumn{1}{c}{---} \\
    \midrule
    \multirow{4}{=}{Cyrillic} & Exact Match & \multicolumn{1}{c}{---} & 0.512 & 0.533 & 0.113 & 0.008 & 0.000 & 0.000 \\
     & Bag of Words & \multicolumn{1}{c}{---} & 0.512 & 0.533 & 0.113 & 0.008 & 0.000 & 0.000 \\
     & $\BLEU$ & \multicolumn{1}{c}{---} & 0.818 & 0.846 & 0.427 & 0.080 & 0.045 & 0.040 \\
     & $\chrFPP$ & \multicolumn{1}{c}{---} & 0.906 & 0.918 & 0.552 & 0.143 & 0.107 & 0.113 \\
    \midrule
    \multirow{4}{=}{Hebrew} & Exact Match & 0.492 & 0.425 & \multicolumn{1}{c}{---} & \multicolumn{1}{c}{---} & \multicolumn{1}{c}{---} & \multicolumn{1}{c}{---} & \multicolumn{1}{c}{---} \\
     & Bag of Words & 0.492 & 0.425 & \multicolumn{1}{c}{---} & \multicolumn{1}{c}{---} & \multicolumn{1}{c}{---} & \multicolumn{1}{c}{---} & \multicolumn{1}{c}{---} \\
     & $\BLEU$ & 0.814 & 0.824 & \multicolumn{1}{c}{---} & \multicolumn{1}{c}{---} & \multicolumn{1}{c}{---} & \multicolumn{1}{c}{---} & \multicolumn{1}{c}{---} \\
     & $\chrFPP$ & 0.948 & 0.920 & \multicolumn{1}{c}{---} & \multicolumn{1}{c}{---} & \multicolumn{1}{c}{---} & \multicolumn{1}{c}{---} & \multicolumn{1}{c}{---} \\
    \midrule
    \multirow{4}{=}{Hebrew + points} & Exact Match & 0.000 & 0.000 & 0.000 & 0.000 & 0.000 & 0.000 & 0.000 \\
     & Bag of Words & 0.000 & 0.000 & 0.000 & 0.000 & 0.000 & 0.000 & 0.000 \\
     & $\BLEU$ & 0.104 & 0.074 & 0.088 & 0.018 & 0.012 & 0.011 & 0.004 \\
     & $\chrFPP$ & 0.456 & 0.386 & 0.386 & 0.128 & 0.061 & 0.039 & 0.025 \\
    \bottomrule
  \end{tabularx}
  \caption{Mean results for \texttt{gpt-5-nano} in the orthography experiment, grouped by target orthography and grammar size.}
  \label{tab:results-orthography-grammar:gpt-5-nano}
\end{table*}

\begin{table*}[ht]
  \centering
  \small
  \sisetup{table-format=1.3}
  \begin{tabularx}{\textwidth}{>{\raggedright\arraybackslash}X l S S S S S S S}
    \toprule
    \textbf{Condition} & \textbf{Metric} & \textbf{46} & \textbf{66} & \textbf{106} & \textbf{826} & \textbf{4,026} & \textbf{6,026} & \textbf{8,026} \\
    \midrule
    \multirow{4}{=}{Latin} & Exact Match & 0.325 & 0.258 & 0.150 & 0.000 & 0.000 & 0.000 & 0.000 \\
     & Bag of Words & 0.325 & 0.263 & 0.150 & 0.000 & 0.000 & 0.000 & 0.000 \\
     & $\BLEU$ & 0.705 & 0.676 & 0.535 & 0.089 & 0.034 & 0.027 & 0.015 \\
     & $\chrFPP$ & 0.818 & 0.799 & 0.685 & 0.263 & 0.161 & 0.138 & 0.106 \\
    \midrule
    \multirow{4}{=}{Latin + diac.} & Exact Match & 0.312 & 0.296 & 0.108 & 0.000 & 0.000 & 0.000 & \multicolumn{1}{c}{---} \\
     & Bag of Words & 0.325 & 0.308 & 0.121 & 0.000 & 0.000 & 0.000 & \multicolumn{1}{c}{---} \\
     & $\BLEU$ & 0.743 & 0.748 & 0.591 & 0.096 & 0.041 & 0.023 & \multicolumn{1}{c}{---} \\
     & $\chrFPP$ & 0.838 & 0.858 & 0.729 & 0.249 & 0.115 & 0.089 & \multicolumn{1}{c}{---} \\
    \midrule
    \multirow{4}{=}{Cyrillic} & Exact Match & 0.308 & 0.287 & 0.192 & 0.000 & 0.000 & 0.000 & \multicolumn{1}{c}{---} \\
     & Bag of Words & 0.396 & 0.300 & 0.204 & 0.000 & 0.000 & 0.000 & \multicolumn{1}{c}{---} \\
     & $\BLEU$ & 0.753 & 0.705 & 0.640 & 0.084 & 0.027 & 0.021 & \multicolumn{1}{c}{---} \\
     & $\chrFPP$ & 0.862 & 0.827 & 0.781 & 0.204 & 0.093 & 0.073 & \multicolumn{1}{c}{---} \\
    \midrule
    \multirow{4}{=}{Hebrew} & Exact Match & 0.433 & 0.338 & 0.237 & 0.000 & 0.000 & 0.000 & \multicolumn{1}{c}{---} \\
     & Bag of Words & 0.446 & 0.371 & 0.246 & 0.000 & 0.000 & 0.000 & \multicolumn{1}{c}{---} \\
     & $\BLEU$ & 0.766 & 0.740 & 0.625 & 0.059 & 0.032 & 0.026 & \multicolumn{1}{c}{---} \\
     & $\chrFPP$ & 0.877 & 0.837 & 0.753 & 0.226 & 0.106 & 0.081 & \multicolumn{1}{c}{---} \\
    \midrule
    \multirow{4}{=}{Hebrew + points} & Exact Match & 0.000 & 0.000 & 0.000 & 0.000 & 0.000 & 0.000 & \multicolumn{1}{c}{---} \\
     & Bag of Words & 0.000 & 0.000 & 0.000 & 0.000 & 0.000 & 0.000 & \multicolumn{1}{c}{---} \\
     & $\BLEU$ & 0.103 & 0.077 & 0.078 & 0.004 & 0.003 & 0.007 & \multicolumn{1}{c}{---} \\
     & $\chrFPP$ & 0.432 & 0.388 & 0.330 & 0.079 & 0.046 & 0.027 & \multicolumn{1}{c}{---} \\
    \bottomrule
  \end{tabularx}
  \caption{Mean results for \texttt{gemma-3-12b-it} in the orthography experiment, grouped by target orthography and grammar size.}
  \label{tab:results-orthography-grammar:google_gemma-3-12b-it}
\end{table*}

\begin{table*}[ht]
  \centering
  \small
  \sisetup{table-format=1.3}
  \begin{tabularx}{\textwidth}{>{\raggedright\arraybackslash}X l S S S S S S S}
    \toprule
    \textbf{Condition} & \textbf{Metric} & \textbf{46} & \textbf{66} & \textbf{106} & \textbf{826} & \textbf{4,026} & \textbf{6,026} & \textbf{8,026} \\
    \midrule
    \multirow{4}{=}{Latin} & Exact Match & 0.004 & 0.000 & 0.004 & 0.000 & 0.000 & 0.000 & 0.000 \\
     & Bag of Words & 0.004 & 0.000 & 0.004 & 0.000 & 0.000 & 0.000 & 0.000 \\
     & $\BLEU$ & 0.016 & 0.017 & 0.026 & 0.003 & 0.002 & 0.001 & 0.002 \\
     & $\chrFPP$ & 0.128 & 0.123 & 0.137 & 0.093 & 0.089 & 0.089 & 0.084 \\
    \midrule
    \multirow{4}{=}{Latin + diac.} & Exact Match & 0.000 & 0.000 & 0.000 & 0.000 & 0.000 & 0.000 & \multicolumn{1}{c}{---} \\
     & Bag of Words & 0.000 & 0.000 & 0.000 & 0.000 & 0.000 & 0.000 & \multicolumn{1}{c}{---} \\
     & $\BLEU$ & 0.008 & 0.012 & 0.007 & 0.000 & 0.000 & 0.000 & \multicolumn{1}{c}{---} \\
     & $\chrFPP$ & 0.057 & 0.074 & 0.056 & 0.041 & 0.039 & 0.035 & \multicolumn{1}{c}{---} \\
    \midrule
    \multirow{4}{=}{Cyrillic} & Exact Match & 0.004 & 0.004 & 0.004 & 0.000 & 0.000 & 0.000 & \multicolumn{1}{c}{---} \\
     & Bag of Words & 0.004 & 0.004 & 0.004 & 0.000 & 0.000 & 0.000 & \multicolumn{1}{c}{---} \\
     & $\BLEU$ & 0.008 & 0.025 & 0.016 & 0.002 & 0.001 & 0.000 & \multicolumn{1}{c}{---} \\
     & $\chrFPP$ & 0.014 & 0.048 & 0.032 & 0.004 & 0.002 & 0.000 & \multicolumn{1}{c}{---} \\
    \midrule
    \multirow{4}{=}{Hebrew} & Exact Match & 0.000 & 0.004 & 0.000 & 0.000 & 0.000 & 0.000 & \multicolumn{1}{c}{---} \\
     & Bag of Words & 0.000 & 0.004 & 0.000 & 0.000 & 0.000 & 0.000 & \multicolumn{1}{c}{---} \\
     & $\BLEU$ & 0.038 & 0.065 & 0.033 & 0.002 & 0.001 & 0.000 & \multicolumn{1}{c}{---} \\
     & $\chrFPP$ & 0.075 & 0.123 & 0.077 & 0.014 & 0.004 & 0.001 & \multicolumn{1}{c}{---} \\
    \midrule
    \multirow{4}{=}{Hebrew + points} & Exact Match & 0.000 & 0.000 & 0.000 & 0.000 & 0.000 & 0.000 & \multicolumn{1}{c}{---} \\
     & Bag of Words & 0.000 & 0.000 & 0.000 & 0.000 & 0.000 & 0.000 & \multicolumn{1}{c}{---} \\
     & $\BLEU$ & 0.006 & 0.004 & 0.011 & 0.000 & 0.000 & 0.000 & \multicolumn{1}{c}{---} \\
     & $\chrFPP$ & 0.028 & 0.023 & 0.044 & 0.004 & 0.001 & 0.002 & \multicolumn{1}{c}{---} \\
    \bottomrule
  \end{tabularx}
  \caption{Mean results for \texttt{gemma-3-4b-it} in the orthography experiment, grouped by target orthography and grammar size.}
  \label{tab:results-orthography-grammar:google_gemma-3-4b-it}
\end{table*}

\begin{table*}[ht]
  \centering
  \small
  \sisetup{table-format=1.3}
  \begin{tabularx}{\textwidth}{>{\raggedright\arraybackslash}X l S S S S S S S}
    \toprule
    \textbf{Condition} & \textbf{Metric} & \textbf{46} & \textbf{66} & \textbf{106} & \textbf{826} & \textbf{4,026} & \textbf{6,026} & \textbf{8,026} \\
    \midrule
    \multirow{4}{=}{Latin} & Exact Match & 0.000 & 0.000 & 0.000 & 0.000 & \multicolumn{1}{c}{---} & \multicolumn{1}{c}{---} & \multicolumn{1}{c}{---} \\
     & Bag of Words & 0.000 & 0.000 & 0.000 & 0.000 & \multicolumn{1}{c}{---} & \multicolumn{1}{c}{---} & \multicolumn{1}{c}{---} \\
     & $\BLEU$ & 0.000 & 0.002 & 0.001 & 0.001 & \multicolumn{1}{c}{---} & \multicolumn{1}{c}{---} & \multicolumn{1}{c}{---} \\
     & $\chrFPP$ & 0.101 & 0.105 & 0.104 & 0.098 & \multicolumn{1}{c}{---} & \multicolumn{1}{c}{---} & \multicolumn{1}{c}{---} \\
    \midrule
    \multirow{4}{=}{Latin + diac.} & Exact Match & 0.000 & 0.000 & 0.000 & 0.000 & \multicolumn{1}{c}{---} & \multicolumn{1}{c}{---} & \multicolumn{1}{c}{---} \\
     & Bag of Words & 0.000 & 0.000 & 0.000 & 0.000 & \multicolumn{1}{c}{---} & \multicolumn{1}{c}{---} & \multicolumn{1}{c}{---} \\
     & $\BLEU$ & 0.000 & 0.000 & 0.000 & 0.000 & \multicolumn{1}{c}{---} & \multicolumn{1}{c}{---} & \multicolumn{1}{c}{---} \\
     & $\chrFPP$ & 0.040 & 0.045 & 0.044 & 0.044 & \multicolumn{1}{c}{---} & \multicolumn{1}{c}{---} & \multicolumn{1}{c}{---} \\
    \midrule
    \multirow{4}{=}{Cyrillic} & Exact Match & 0.000 & 0.000 & 0.000 & 0.000 & \multicolumn{1}{c}{---} & \multicolumn{1}{c}{---} & \multicolumn{1}{c}{---} \\
     & Bag of Words & 0.000 & 0.000 & 0.000 & 0.000 & \multicolumn{1}{c}{---} & \multicolumn{1}{c}{---} & \multicolumn{1}{c}{---} \\
     & $\BLEU$ & 0.000 & 0.000 & 0.000 & 0.000 & \multicolumn{1}{c}{---} & \multicolumn{1}{c}{---} & \multicolumn{1}{c}{---} \\
     & $\chrFPP$ & 0.000 & 0.000 & 0.000 & 0.000 & \multicolumn{1}{c}{---} & \multicolumn{1}{c}{---} & \multicolumn{1}{c}{---} \\
    \midrule
    \multirow{4}{=}{Hebrew} & Exact Match & 0.000 & 0.000 & 0.000 & 0.000 & \multicolumn{1}{c}{---} & \multicolumn{1}{c}{---} & \multicolumn{1}{c}{---} \\
     & Bag of Words & 0.000 & 0.000 & 0.000 & 0.000 & \multicolumn{1}{c}{---} & \multicolumn{1}{c}{---} & \multicolumn{1}{c}{---} \\
     & $\BLEU$ & 0.000 & 0.000 & 0.000 & 0.000 & \multicolumn{1}{c}{---} & \multicolumn{1}{c}{---} & \multicolumn{1}{c}{---} \\
     & $\chrFPP$ & 0.000 & 0.000 & 0.000 & 0.000 & \multicolumn{1}{c}{---} & \multicolumn{1}{c}{---} & \multicolumn{1}{c}{---} \\
    \midrule
    \multirow{4}{=}{Hebrew + points} & Exact Match & 0.000 & 0.000 & 0.000 & 0.000 & \multicolumn{1}{c}{---} & \multicolumn{1}{c}{---} & \multicolumn{1}{c}{---} \\
     & Bag of Words & 0.000 & 0.000 & 0.000 & 0.000 & \multicolumn{1}{c}{---} & \multicolumn{1}{c}{---} & \multicolumn{1}{c}{---} \\
     & $\BLEU$ & 0.000 & 0.000 & 0.000 & 0.000 & \multicolumn{1}{c}{---} & \multicolumn{1}{c}{---} & \multicolumn{1}{c}{---} \\
     & $\chrFPP$ & 0.000 & 0.000 & 0.000 & 0.000 & \multicolumn{1}{c}{---} & \multicolumn{1}{c}{---} & \multicolumn{1}{c}{---} \\
    \bottomrule
  \end{tabularx}
  \caption{Mean results for \texttt{gemma-3-1b-it} in the orthography experiment, grouped by target orthography and grammar size.}
  \label{tab:results-orthography-grammar:google_gemma-3-1b-it}
\end{table*}
\begin{table*}[ht]
  \centering
  \small
  \sisetup{table-format=1.3}
  \begin{tabularx}{\textwidth}{>{\raggedright\arraybackslash}X l S S S S S}
    \toprule
    \textbf{Condition} & \textbf{Metric} & \textbf{6.5} & \textbf{11} & \textbf{15.5} & \textbf{20.5} & \textbf{34} \\
    \midrule
    \multirow{4}{=}{Latin} & Exact Match & 0.796 & 0.706 & 0.685 & 0.635 & 0.631 \\
     & Bag of Words & 0.796 & 0.706 & 0.685 & 0.635 & 0.631 \\
     & $\BLEU$ & 0.907 & 0.899 & 0.912 & 0.898 & 0.907 \\
     & $\chrFPP$ & 0.944 & 0.943 & 0.951 & 0.942 & 0.948 \\
    \midrule
    \multirow{4}{=}{Latin + diac.} & Exact Match & 0.806 & 0.696 & 0.696 & 0.653 & 0.564 \\
     & Bag of Words & 0.806 & 0.696 & 0.696 & 0.653 & 0.564 \\
     & $\BLEU$ & 0.920 & 0.901 & 0.914 & 0.906 & 0.880 \\
     & $\chrFPP$ & 0.944 & 0.936 & 0.946 & 0.942 & 0.922 \\
    \midrule
    \multirow{4}{=}{Cyrillic} & Exact Match & 0.754 & 0.680 & 0.636 & 0.522 & 0.542 \\
     & Bag of Words & 0.754 & 0.680 & 0.636 & 0.522 & 0.542 \\
     & $\BLEU$ & 0.883 & 0.880 & 0.884 & 0.854 & 0.856 \\
     & $\chrFPP$ & 0.934 & 0.932 & 0.930 & 0.914 & 0.914 \\
    \midrule
    \multirow{4}{=}{Hebrew} & Exact Match & 0.778 & 0.691 & 0.642 & 0.556 & 0.514 \\
     & Bag of Words & 0.778 & 0.691 & 0.642 & 0.556 & 0.514 \\
     & $\BLEU$ & 0.894 & 0.878 & 0.870 & 0.862 & 0.841 \\
     & $\chrFPP$ & 0.934 & 0.927 & 0.920 & 0.913 & 0.903 \\
    \midrule
    \multirow{4}{=}{Hebrew + points} & Exact Match & 0.000 & 0.000 & 0.000 & 0.000 & 0.000 \\
     & Bag of Words & 0.000 & 0.000 & 0.000 & 0.000 & 0.000 \\
     & $\BLEU$ & 0.117 & 0.083 & 0.070 & 0.074 & 0.066 \\
     & $\chrFPP$ & 0.408 & 0.410 & 0.404 & 0.407 & 0.408 \\
    \bottomrule
  \end{tabularx}
  \caption{Mean results for \texttt{gpt-5} in the orthography experiment, grouped by target orthography and input string length.}
  \label{tab:results-orthography-length:gpt-5}
\end{table*}

\begin{table*}[ht]
  \centering
  \small
  \sisetup{table-format=1.3}
  \begin{tabularx}{\textwidth}{>{\raggedright\arraybackslash}X l S S S S S}
    \toprule
    \textbf{Condition} & \textbf{Metric} & \textbf{6.5} & \textbf{11} & \textbf{15.5} & \textbf{20.5} & \textbf{34} \\
    \midrule
    \multirow{4}{=}{Latin} & Exact Match & 0.580 & 0.555 & 0.500 & 0.472 & 0.521 \\
     & Bag of Words & 0.580 & 0.555 & 0.500 & 0.472 & 0.521 \\
     & $\BLEU$ & 0.746 & 0.748 & 0.716 & 0.712 & 0.730 \\
     & $\chrFPP$ & 0.823 & 0.842 & 0.821 & 0.816 & 0.821 \\
    \midrule
    \multirow{4}{=}{Latin + diac.} & Exact Match & 0.633 & 0.529 & 0.513 & 0.532 & 0.412 \\
     & Bag of Words & 0.633 & 0.529 & 0.513 & 0.532 & 0.412 \\
     & $\BLEU$ & 0.781 & 0.740 & 0.744 & 0.755 & 0.666 \\
     & $\chrFPP$ & 0.841 & 0.821 & 0.835 & 0.832 & 0.759 \\
    \midrule
    \multirow{4}{=}{Cyrillic} & Exact Match & 0.585 & 0.538 & 0.543 & 0.423 & 0.479 \\
     & Bag of Words & 0.585 & 0.538 & 0.543 & 0.423 & 0.479 \\
     & $\BLEU$ & 0.739 & 0.745 & 0.742 & 0.664 & 0.707 \\
     & $\chrFPP$ & 0.821 & 0.829 & 0.830 & 0.769 & 0.798 \\
    \midrule
    \multirow{4}{=}{Hebrew} & Exact Match & 0.551 & 0.386 & 0.409 & 0.407 & 0.334 \\
     & Bag of Words & 0.551 & 0.386 & 0.409 & 0.407 & 0.334 \\
     & $\BLEU$ & 0.721 & 0.659 & 0.667 & 0.661 & 0.640 \\
     & $\chrFPP$ & 0.800 & 0.788 & 0.779 & 0.775 & 0.742 \\
    \midrule
    \multirow{4}{=}{Hebrew + points} & Exact Match & 0.000 & 0.000 & 0.000 & 0.000 & 0.000 \\
     & Bag of Words & 0.000 & 0.000 & 0.000 & 0.000 & 0.000 \\
     & $\BLEU$ & 0.104 & 0.078 & 0.062 & 0.067 & 0.055 \\
     & $\chrFPP$ & 0.365 & 0.376 & 0.359 & 0.365 & 0.347 \\
    \bottomrule
  \end{tabularx}
  \caption{Mean results for \texttt{gpt-5-mini} in the orthography experiment, grouped by target orthography and input string length.}
  \label{tab:results-orthography-length:gpt-5-mini}
\end{table*}

\begin{table*}[ht]
  \centering
  \small
  \sisetup{table-format=1.3}
  \begin{tabularx}{\textwidth}{>{\raggedright\arraybackslash}X l S S S S S}
    \toprule
    \textbf{Condition} & \textbf{Metric} & \textbf{6.5} & \textbf{11} & \textbf{15.5} & \textbf{20.5} & \textbf{34} \\
    \midrule
    \multirow{4}{=}{Latin} & Exact Match & 0.016 & 0.000 & 0.000 & 0.000 & 0.000 \\
     & Bag of Words & 0.016 & 0.000 & 0.000 & 0.000 & 0.000 \\
     & $\BLEU$ & 0.168 & 0.069 & 0.040 & 0.020 & 0.011 \\
     & $\chrFPP$ & 0.327 & 0.223 & 0.165 & 0.140 & 0.110 \\
    \midrule
    \multirow{4}{=}{Latin + diac.} & Exact Match & 0.786 & 0.603 & 0.595 & 0.435 & 0.429 \\
     & Bag of Words & 0.786 & 0.603 & 0.595 & 0.435 & 0.429 \\
     & $\BLEU$ & 0.918 & 0.878 & 0.885 & 0.881 & 0.829 \\
     & $\chrFPP$ & 0.954 & 0.946 & 0.946 & 0.941 & 0.893 \\
    \midrule
    \multirow{4}{=}{Cyrillic} & Exact Match & 0.246 & 0.257 & 0.186 & 0.144 & 0.140 \\
     & Bag of Words & 0.246 & 0.257 & 0.186 & 0.144 & 0.140 \\
     & $\BLEU$ & 0.435 & 0.391 & 0.381 & 0.324 & 0.341 \\
     & $\chrFPP$ & 0.545 & 0.482 & 0.448 & 0.391 & 0.402 \\
    \midrule
    \multirow{4}{=}{Hebrew} & Exact Match & 0.691 & 0.430 & 0.418 & 0.350 & 0.300 \\
     & Bag of Words & 0.691 & 0.430 & 0.418 & 0.350 & 0.300 \\
     & $\BLEU$ & 0.855 & 0.827 & 0.787 & 0.808 & 0.805 \\
     & $\chrFPP$ & 0.951 & 0.946 & 0.917 & 0.930 & 0.917 \\
    \midrule
    \multirow{4}{=}{Hebrew + points} & Exact Match & 0.000 & 0.000 & 0.000 & 0.000 & 0.000 \\
     & Bag of Words & 0.000 & 0.000 & 0.000 & 0.000 & 0.000 \\
     & $\BLEU$ & 0.070 & 0.049 & 0.035 & 0.038 & 0.026 \\
     & $\chrFPP$ & 0.255 & 0.248 & 0.199 & 0.201 & 0.155 \\
    \bottomrule
  \end{tabularx}
  \caption{Mean results for \texttt{gpt-5-nano} in the orthography experiment, grouped by target orthography and input string length.}
  \label{tab:results-orthography-length:gpt-5-nano}
\end{table*}

\begin{table*}[ht]
  \centering
  \small
  \sisetup{table-format=1.3}
  \begin{tabularx}{\textwidth}{>{\raggedright\arraybackslash}X l S S S S S}
    \toprule
    \textbf{Condition} & \textbf{Metric} & \textbf{6.5} & \textbf{11} & \textbf{15.5} & \textbf{20.5} & \textbf{34} \\
    \midrule
    \multirow{4}{=}{Latin} & Exact Match & 0.168 & 0.136 & 0.062 & 0.078 & 0.074 \\
     & Bag of Words & 0.168 & 0.136 & 0.062 & 0.081 & 0.074 \\
     & $\BLEU$ & 0.318 & 0.331 & 0.270 & 0.291 & 0.280 \\
     & $\chrFPP$ & 0.426 & 0.452 & 0.405 & 0.428 & 0.417 \\
    \midrule
    \multirow{4}{=}{Latin + diac.} & Exact Match & 0.190 & 0.127 & 0.104 & 0.077 & 0.081 \\
     & Bag of Words & 0.199 & 0.139 & 0.104 & 0.083 & 0.086 \\
     & $\BLEU$ & 0.377 & 0.377 & 0.375 & 0.395 & 0.327 \\
     & $\chrFPP$ & 0.457 & 0.494 & 0.484 & 0.513 & 0.445 \\
    \midrule
    \multirow{4}{=}{Cyrillic} & Exact Match & 0.189 & 0.148 & 0.152 & 0.083 & 0.073 \\
     & Bag of Words & 0.206 & 0.181 & 0.168 & 0.094 & 0.091 \\
     & $\BLEU$ & 0.358 & 0.404 & 0.395 & 0.354 & 0.354 \\
     & $\chrFPP$ & 0.445 & 0.507 & 0.499 & 0.461 & 0.465 \\
    \midrule
    \multirow{4}{=}{Hebrew} & Exact Match & 0.262 & 0.198 & 0.161 & 0.126 & 0.078 \\
     & Bag of Words & 0.271 & 0.210 & 0.172 & 0.139 & 0.078 \\
     & $\BLEU$ & 0.411 & 0.385 & 0.377 & 0.354 & 0.341 \\
     & $\chrFPP$ & 0.506 & 0.494 & 0.486 & 0.459 & 0.455 \\
    \midrule
    \multirow{4}{=}{Hebrew + points} & Exact Match & 0.000 & 0.000 & 0.000 & 0.000 & 0.000 \\
     & Bag of Words & 0.000 & 0.000 & 0.000 & 0.000 & 0.000 \\
     & $\BLEU$ & 0.065 & 0.044 & 0.038 & 0.045 & 0.028 \\
     & $\chrFPP$ & 0.226 & 0.225 & 0.219 & 0.225 & 0.191 \\
    \bottomrule
  \end{tabularx}
  \caption{Mean results for \texttt{gemma-3-12b-it} in the orthography experiment, grouped by target orthography and input string length.}
  \label{tab:results-orthography-length:google_gemma-3-12b-it}
\end{table*}

\begin{table*}[ht]
  \centering
  \small
  \sisetup{table-format=1.3}
  \begin{tabularx}{\textwidth}{>{\raggedright\arraybackslash}X l S S S S S}
    \toprule
    \textbf{Condition} & \textbf{Metric} & \textbf{6.5} & \textbf{11} & \textbf{15.5} & \textbf{20.5} & \textbf{34} \\
    \midrule
    \multirow{4}{=}{Latin} & Exact Match & 0.000 & 0.007 & 0.000 & 0.000 & 0.000 \\
     & Bag of Words & 0.000 & 0.007 & 0.000 & 0.000 & 0.000 \\
     & $\BLEU$ & 0.010 & 0.021 & 0.007 & 0.009 & 0.004 \\
     & $\chrFPP$ & 0.092 & 0.118 & 0.105 & 0.112 & 0.107 \\
    \midrule
    \multirow{4}{=}{Latin + diac.} & Exact Match & 0.000 & 0.000 & 0.000 & 0.000 & 0.000 \\
     & Bag of Words & 0.000 & 0.000 & 0.000 & 0.000 & 0.000 \\
     & $\BLEU$ & 0.005 & 0.004 & 0.001 & 0.006 & 0.008 \\
     & $\chrFPP$ & 0.042 & 0.049 & 0.047 & 0.060 & 0.057 \\
    \midrule
    \multirow{4}{=}{Cyrillic} & Exact Match & 0.006 & 0.000 & 0.003 & 0.000 & 0.000 \\
     & Bag of Words & 0.006 & 0.000 & 0.003 & 0.000 & 0.000 \\
     & $\BLEU$ & 0.014 & 0.011 & 0.009 & 0.003 & 0.005 \\
     & $\chrFPP$ & 0.021 & 0.023 & 0.018 & 0.010 & 0.011 \\
    \midrule
    \multirow{4}{=}{Hebrew} & Exact Match & 0.003 & 0.000 & 0.000 & 0.000 & 0.000 \\
     & Bag of Words & 0.003 & 0.000 & 0.000 & 0.000 & 0.000 \\
     & $\BLEU$ & 0.032 & 0.041 & 0.020 & 0.013 & 0.011 \\
     & $\chrFPP$ & 0.058 & 0.079 & 0.048 & 0.033 & 0.030 \\
    \midrule
    \multirow{4}{=}{Hebrew + points} & Exact Match & 0.000 & 0.000 & 0.000 & 0.000 & 0.000 \\
     & Bag of Words & 0.000 & 0.000 & 0.000 & 0.000 & 0.000 \\
     & $\BLEU$ & 0.005 & 0.004 & 0.005 & 0.002 & 0.002 \\
     & $\chrFPP$ & 0.021 & 0.018 & 0.023 & 0.012 & 0.012 \\
    \bottomrule
  \end{tabularx}
  \caption{Mean results for \texttt{gemma-3-4b-it} in the orthography experiment, grouped by target orthography and input string length.}
  \label{tab:results-orthography-length:google_gemma-3-4b-it}
\end{table*}

\begin{table*}[ht]
  \centering
  \small
  \sisetup{table-format=1.3}
  \begin{tabularx}{\textwidth}{>{\raggedright\arraybackslash}X l S S S S S}
    \toprule
    \textbf{Condition} & \textbf{Metric} & \textbf{6.5} & \textbf{11} & \textbf{15.5} & \textbf{20.5} & \textbf{34} \\
    \midrule
    \multirow{4}{=}{Latin} & Exact Match & 0.000 & 0.000 & 0.000 & 0.000 & 0.000 \\
     & Bag of Words & 0.000 & 0.000 & 0.000 & 0.000 & 0.000 \\
     & $\BLEU$ & 0.001 & 0.001 & 0.000 & 0.001 & 0.001 \\
     & $\chrFPP$ & 0.084 & 0.100 & 0.104 & 0.110 & 0.116 \\
    \midrule
    \multirow{4}{=}{Latin + diac.} & Exact Match & 0.000 & 0.000 & 0.000 & 0.000 & 0.000 \\
     & Bag of Words & 0.000 & 0.000 & 0.000 & 0.000 & 0.000 \\
     & $\BLEU$ & 0.000 & 0.000 & 0.000 & 0.000 & 0.000 \\
     & $\chrFPP$ & 0.035 & 0.042 & 0.046 & 0.047 & 0.048 \\
    \midrule
    \multirow{4}{=}{Cyrillic} & Exact Match & 0.000 & 0.000 & 0.000 & 0.000 & 0.000 \\
     & Bag of Words & 0.000 & 0.000 & 0.000 & 0.000 & 0.000 \\
     & $\BLEU$ & 0.000 & 0.000 & 0.000 & 0.000 & 0.000 \\
     & $\chrFPP$ & 0.000 & 0.000 & 0.000 & 0.000 & 0.000 \\
    \midrule
    \multirow{4}{=}{Hebrew} & Exact Match & 0.000 & 0.000 & 0.000 & 0.000 & 0.000 \\
     & Bag of Words & 0.000 & 0.000 & 0.000 & 0.000 & 0.000 \\
     & $\BLEU$ & 0.000 & 0.000 & 0.000 & 0.000 & 0.000 \\
     & $\chrFPP$ & 0.000 & 0.000 & 0.000 & 0.000 & 0.000 \\
    \midrule
    \multirow{4}{=}{Hebrew + points} & Exact Match & 0.000 & 0.000 & 0.000 & 0.000 & 0.000 \\
     & Bag of Words & 0.000 & 0.000 & 0.000 & 0.000 & 0.000 \\
     & $\BLEU$ & 0.000 & 0.000 & 0.000 & 0.000 & 0.000 \\
     & $\chrFPP$ & 0.000 & 0.000 & 0.000 & 0.000 & 0.000 \\
    \bottomrule
  \end{tabularx}
  \caption{Mean results for \texttt{gemma-3-1b-it} in the orthography experiment, grouped by target orthography and input string length.}
  \label{tab:results-orthography-length:google_gemma-3-1b-it}
\end{table*}

\end{document}